\documentclass[sigconf, nonacm]{acmart}
\usepackage{amsmath,bm}
\usepackage{booktabs}
\usepackage{multirow}
\usepackage{graphicx}
\usepackage{pifont}
\usepackage{enumitem}
\usepackage{siunitx}
\usepackage{xspace}
\usepackage{array}
\usepackage{balance}
\usepackage{ulem}
\usepackage[utf8]{inputenc}
\AtBeginDocument{%
  }

\newcommand{\cmark}{$\surd$}
\newcommand{\xmark}{$\times$}
\newcommand{\R}{\mathbb{R}}
\newcommand{\vect}[1]{\mathbf{#1}}
\newcommand{\norm}[1]{\left\lVert #1 \right\rVert}

\acmSubmissionID{1562}

\author{
Jingzhou Luo$^1$, \space Yifan Wen$^1$, \space Yongjie Bai$^{1,2}$, \space Xinshuai Song$^1$, \space Yang Liu$^{1,3,*}$, \space Liang Lin$^{1,2,3,4}$ \\
\small $^1$Sun Yat-sen University, China \space  $^2$Peng Cheng Laboratory \space $^3$Guangdong Key Laboratory of Big Data Analysis and Processing \space $^4$X-Era AI Lab \\
\small *Corresponding Author
}
\begin{document}

% \title{The Name of the Title Is Hope}
% \title[RoVLA]{RoVLA: A Multi-aspect Transformation Consistency Constrained Framework for Robust Visual-Language-Action}
% \title{RoVLA: Multi-aspect Consistency Constraints for Robust Vision-Language-Action Learning}

\title{RoVLA: Multi-Consistency Constraints for Robust Vision-Language-Action Models}

\renewcommand{\shortauthors}{Jingzhou Luo et al.}

\begin{abstract}
Vision-Language-Action (VLA) models have shown strong performance on embodied manipulation, yet they remain brittle under visual observation changes, paraphrased language instructions, and compounded perturbations. This limitation suggests that existing methods still rely heavily on shallow correlations in the training distribution, rather than learning stable couplings among task semantics, environment states, and action generation. Although recent efforts improve robustness through larger-scale training, post-training adaptation, or enhanced predictive modeling, they rarely enforce invariance-oriented consistency within the end-to-end policy itself.
To address this issue, we propose RoVLA, a robust vision-language-action framework with multi-consistency constraints. RoVLA enforces consistency under three complementary transformations: instruction semantics, trajectory evolution, and observation perturbation. Specifically, Instructional Consistency (IC) promotes stable grounding under semantically equivalent instruction rewrites, Evolutionary Consistency (EC) preserves coherent action intent throughout the generation process, and Observational Consistency (OC) improves robustness to visual and proprioceptive perturbations by enforcing consistent predictions before and after targeted disturbances. By explicitly modeling these invariances during training, RoVLA reduces reliance on superficial correlations and improves robustness and generalization.
Experiments on LIBERO-Plus, RoboTwin 2.0, and real-world manipulation tasks show that RoVLA consistently outperforms strong baseline methods and exhibits superior robustness under diverse task and observation shifts. These results demonstrate the effectiveness of multi-consistency learning for robust embodied control. Codes will be available at \href{https://github.com/HCPLab-SYSU/RoVLA}{https://github.com/HCPLab-SYSU/RoVLA}.
\end{abstract}




% %% A "teaser" image appears between the author and affiliation
% %% information and the body of the document, and typically spans the


\maketitle

\begin{figure}[t]
    \centering
    \includegraphics[width=1.0\linewidth]{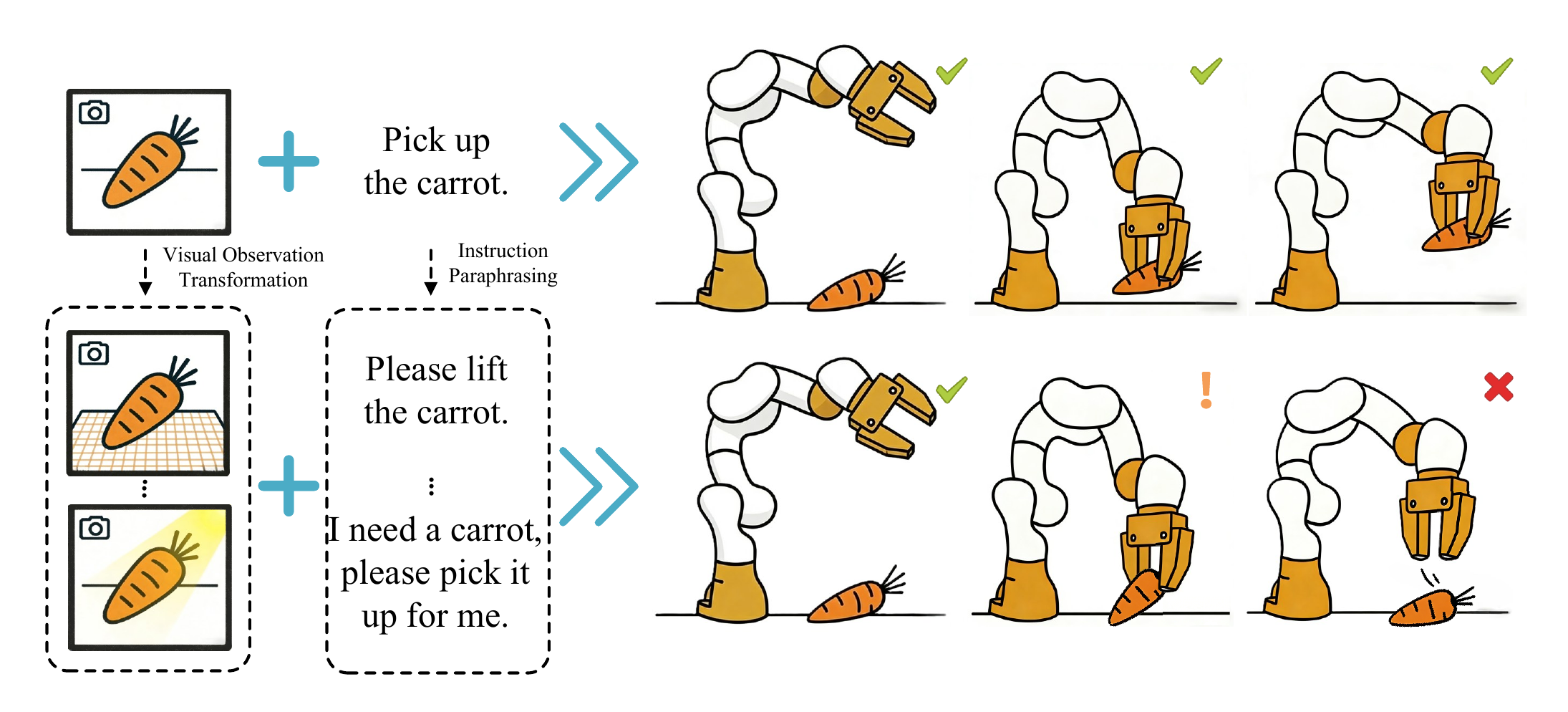}
    \vspace{-15pt}
    \caption{Existing VLA models often lack robustness under visual observation changes and paraphrased but semantically equivalent language instructions.}
    \vspace{-5pt}
    \label{fig:RoVLA-motivation}
\end{figure}

\section{Introduction}

Vision-Language-Action (VLA) modeling has emerged as a promising paradigm for instruction-following robotic assistance in the physical world, integrating visual perception~\cite{wang2025robot, dosovitskiy2021vit,qi2017pointnet++,luo2025dspnet}, language understanding~\cite{aghzal2025survey, yang2025qwen3, wang2025internvl3_5}, and action generation~\cite{peebles2023scalable, lipman2022flow, chi2025diffusion} into a unified end-to-end policy~\cite{zhong2025survey, kim2024openvla, black2024pi_0, chen2025internvla, bjorck2025gr00t}. Recent methods, including the $\pi$ family~\cite{black2024pi_0,black2025pi_05}, the GR00T family~\cite{bjorck2025gr00t,nvidia2025gr00t}, the RT family~\cite{brohan2022rt,zitkovich2023rt}, and OpenVLA~\cite{kim2024openvla}, have shown that large pretrained multimodal models can be effectively adapted to robotic manipulation and achieve strong performance on standard benchmarks. However, high average performance under matched training and testing conditions does not necessarily translate into robust embodied competence in open environments~\cite{fei2025libero, chen2025robotwin, wang2025exploring, yang2025instructvla}. Fei et al.~\cite{fei2025libero} further identify three representative failure modes of existing VLA models on LIBERO-Plus: pseudo visual understanding, where policies are highly sensitive to viewpoint, background, or illumination changes; pseudo language following, where semantically equivalent instructions induce inconsistent actions; and pseudo compositional generalization, where multiple mild perturbations jointly cause severe degradation. These results indicate that current VLA models still struggle to maintain robust perception, language grounding, and action execution beyond the training distribution.
As illustrated in Fig.~\ref{fig:RoVLA-motivation}, existing VLA models may succeed when visual observations and language instructions closely match training conditions, yet fail under slight scene changes or semantically equivalent instruction reformulations~\cite{fei2025libero}. This suggests that many current policies still rely on shallow cross-modal correlations rather than stable task-relevant invariances.

From this perspective, the central limitation of current VLA models lies not primarily in insufficient capacity within any single modality, but in inadequate invariant policy learning across task-relevant transformations. At the language level, models often fail to form consistent action representations for semantically equivalent instructions. At the action-generation level, the representation of the same task intent may drift across different evolution stages. At the observation level, policies remain vulnerable to visual changes and proprioceptive deviations. Without explicit modeling of these invariances, VLA policies tend to overfit superficial training statistics and generalize poorly in complex open environments.

% Existing efforts to improve VLA robustness mainly follow broader-data or stronger-adaptation routes. Co-training strategies such as $\pi_{0.5}$~\cite{black2025pi_05} expand supervision with heterogeneous robot data and richer semantic annotations. Post-training approaches such as RIPT-VLA~\cite{tan2025ript} further adapt policies through reinforcement learning. World-model-based methods, including WorldVLA~\cite{cen2025worldvla} and UniVLA~\cite{wang2025univla}, strengthen predictive modeling and long-horizon planning. While effective, these methods largely treat robustness as a byproduct of larger-scale data, stronger optimization, or improved dynamics modeling, rather than as an explicit objective of invariant policy learning.

Existing efforts to improve VLA robustness mainly pursue two directions: scaling up training data and strengthening post hoc adaptation. Co-training strategies, such as 
$\pi_{0.5}$~\cite{black2025pi_05}, expand supervision through heterogeneous robot data and richer semantic annotations. Post-training methods, such as RIPT-VLA~\cite{tan2025ript}, further enhance policy performance through reinforcement learning. In parallel, world-model-based approaches, including WorldVLA~\cite{cen2025worldvla} and UniVLA~\cite{wang2025univla}, improve predictive modeling and long-horizon planning. Although effective, these methods largely treat robustness as an emergent benefit of larger-scale data, stronger optimization, or improved dynamics modeling, rather than as an explicit objective of invariant policy learning.
% To address these issues, we propose RoVLA (\textbf{Ro}bust \textbf{V}ision-\textbf{L}anguage-\textbf{A}ction), a Multi-Consistency Constrained framework for robust VLA learning. RoVLA explicitly constrains the policy along three transformation aspects that commonly induce VLA failures: instruction semantic transformation, trajectory evolution transformation, and observation perturbation transformation. Specifically, it introduces Instructional Consistency (IC) to encourage stable grounding under semantically equivalent instruction reformulations, Evolutionary Consistency (EC) to preserve task intent across different action evolution stages, and Observational Consistency (OC) to enforce prediction consistency under targeted observation perturbations, thereby improving robustness to viewpoint shifts, background variation, illumination changes, and robot-state deviations.

% Compared with methods that rely primarily on broader training exposure or post-hoc adaptation, RoVLA directly incorporates robustness priors into end-to-end policy learning. The three consistency constraints operate at complementary levels, namely semantic input, generative evolution, and observation robustness, forming a unified Multi-Consistency constrained framework for robust VLA modeling. By explicitly promoting invariant task-semantic representations and stable action behaviors under diverse transformations, RoVLA reduces dependence on superficial correlations and improves both robustness and generalization in dynamic environments.

To address these challenges, we propose \textbf{RoVLA} (\textbf{Ro}bust \textbf{V}ision-\textbf{L}anguage-\textbf{A}ction), a multi-consistency constrained framework for robust VLA learning. RoVLA explicitly constrains the policy under three transformations that commonly induce failure in VLA models: instruction semantic transformation, trajectory evolution transformation, and observation perturbation transformation. Specifically, it introduces Instructional Consistency (IC) to promote stable grounding under semantically equivalent instruction reformulations, Evolutionary Consistency (EC) to preserve task intent across different action evolution stages, and Observational Consistency (OC) to enforce prediction stability under targeted observation perturbations. By integrating these complementary constraints into end-to-end policy learning, RoVLA reduces reliance on superficial correlations and improves both robustness and generalization under
distribution shifts.

We evaluate RoVLA on LIBERO-Plus~\cite{fei2025libero}, RoboTwin 2.0~\cite{chen2025robotwin}, and real-world tasks. Experimental results show that RoVLA consistently outperforms existing baselines and exhibits stronger robustness and generalization under diverse perturbations and task variations. Our main contributions are summarized as follows:
\begin{enumerate}
    \item To improve the robustness and generalization of VLA models, we propose RoVLA, a robust VLA framework that explicitly enforces invariant policy learning under three task-preserving transformations: instruction reformulation, trajectory evolution, and observation perturbation. 

\item We introduce three complementary consistency mechanisms for end-to-end VLA training: Instructional Consistency (IC) for language invariance, Evolutionary Consistency (EC) for stable action-intent modeling across flow-matching stages, and Observational Consistency (OC) for robustness to visual and proprioceptive disturbances.
    % \item We introduce Instructional Consistency (IC), which constructs semantically equivalent paraphrase sets and randomly samples paraphrased instructions during training to improve generalization under linguistic variation.

    % \item We introduce Evolutionary Consistency (EC) to improve action generation stability by enforcing consistent predictions across evolution stages, coupled with Observational Consistency (OC) to enhance observational robustness by maintaining prediction consistency under observation perturbations.
    
    \item Extensive experiments on LIBERO-Plus, RoboTwin 2.0, and real-world manipulation tasks show that RoVLA consistently achieves the best overall performance and stronger robustness under diverse task and observation shifts.
\end{enumerate}

% \vspace{-5pt}
\section{Related Work}

\subsection{Generalist VLA Models}
Recent progress in embodied policy learning has been driven by generalist robot policies that jointly model visual observations, language instructions, and motor actions within a unified framework~\cite{liu2025aligning}. Early large-scale systems such as RT-1~\cite{brohan2022rt} and RT-2~\cite{zitkovich2023rt} showed that scaling transformer-based policies with diverse robot data and pretrained vision-language priors can substantially improve multi-task manipulation performance and generalization. Open-source efforts further advanced this direction. Open X-Embodiment~\cite{o2024open} and Octo~\cite{team2024octo} promoted cross-embodiment policy pretraining on large heterogeneous robot datasets, while OpenVLA~\cite{kim2024openvla} established a strong open-source VLA baseline that supports effective downstream fine-tuning. GO-1~\cite{bu2025agibot} further explores scalable generalist policy learning by introducing latent action representations for heterogeneous robot data, demonstrating strong transferability in real-world manipulation. More recent developments, such as the $\pi$ family~\cite{black2024pi_0,black2025pi_05, intelligence2025pi06} and the GR00T family~\cite{bjorck2025gr00t,nvidia2025gr00t}, further advance general robot manipulation through stronger continuous control and dual-system VLA architectures.

In parallel, action generation in embodied policies has evolved from discrete token prediction to continuous generative modeling. Diffusion Policy~\cite{chi2025diffusion} demonstrated the effectiveness of diffusion-based action generation for visuomotor control. Building on this trend, recent VLA methods increasingly adopt continuous action generation to improve precision and efficiency. For example, $\pi_0$~\cite{black2024pi_0} formulates action prediction with flow matching, OpenVLA-OFT~\cite{kim2025openvlaoft} replaces autoregressive decoding with parallel continuous regression, and $\pi_0$-Fast~\cite{pertsch2025fast} improves autoregressive VLA scaling via compressed action tokenization. Beyond standalone policy learning, UniVLA~\cite{wang2025univla} and WorldVLA~\cite{cen2025worldvla} further integrate policy learning with unified token modeling or predictive world modeling to improve scalability and long-horizon reasoning. Overall, existing VLA research has substantially advanced policy capacity, data scalability, and action generation quality, but robustness to task-preserving transformations remains comparatively underexplored.

\subsection{VLA Robustness}
As VLA models achieve increasingly strong results on standard benchmarks, robustness under distribution shift has become a central challenge. LIBERO-Plus~\cite{fei2025libero} systematically reveals that many VLA policies remain fragile under perturbations in language instructions, camera viewpoints, lighting, layout, background, sensor noise, and robot initial states. Similar findings are reported by VLATest~\cite{wang2025vlatest}, which evaluates robustness under scene fuzzing, confounding objects, unseen objects, camera pose variations, lighting changes, and instruction mutations. Collectively, these studies highlight that high average performance under matched train--test conditions does not necessarily translate into reliable embodied robustness in open environments.

Existing efforts to improve VLA robustness mainly follow three directions. The first expands training exposure through heterogeneous robot data, richer semantic supervision, or cross-domain co-training, as represented by $\pi_{0.5}$~\cite{black2025pi_05}. The second improves robustness indirectly through stronger generative or predictive modeling, including unified world modeling and multimodal priors, as in UniVLA~\cite{wang2025univla}, WorldVLA~\cite{cen2025worldvla}, RynnVLA~\cite{cen2025rynnvla}, and Motus~\cite{bi2025motus}. The third adapts pretrained VLA policies via post-training interaction or reinforcement learning, such as RIPT-VLA~\cite{tan2025ript}, VLA-RL~\cite{lu2025vla}, SimpleVLA-RL~\cite{li2025simplevla}, and RobustVLA~\cite{zhang2025robustvla}. Although effective, these approaches mostly treat robustness as a byproduct of broader data, stronger priors, or downstream adaptation, rather than an explicit objective during end-to-end policy learning.
Beyond these VLA-specific efforts, consistency learning provides a complementary perspective on robustness by enforcing prediction stability under semantics-preserving transformations. Representative methods include Mean Teacher~\cite{tarvainen2017meanteacher}, VAT~\cite{miyato2018virtual}, and FixMatch~\cite{sohn2020fixmatch}, while DrQ~\cite{yarats2021drq} shows the effectiveness of consistency regularization for pixel-based control in visual reinforcement learning. However, these methods are not designed for end-to-end VLA policies, where robustness depends on coupled invariances across language grounding, observation perturbations, and action generation. As a result, policy learning under explicit multi-consistency constraints remains insufficiently explored in current VLA research.

% \yongjie{In recent years, the }

\begin{figure*}[t]
    \centering
    \includegraphics[width=0.9\textwidth]{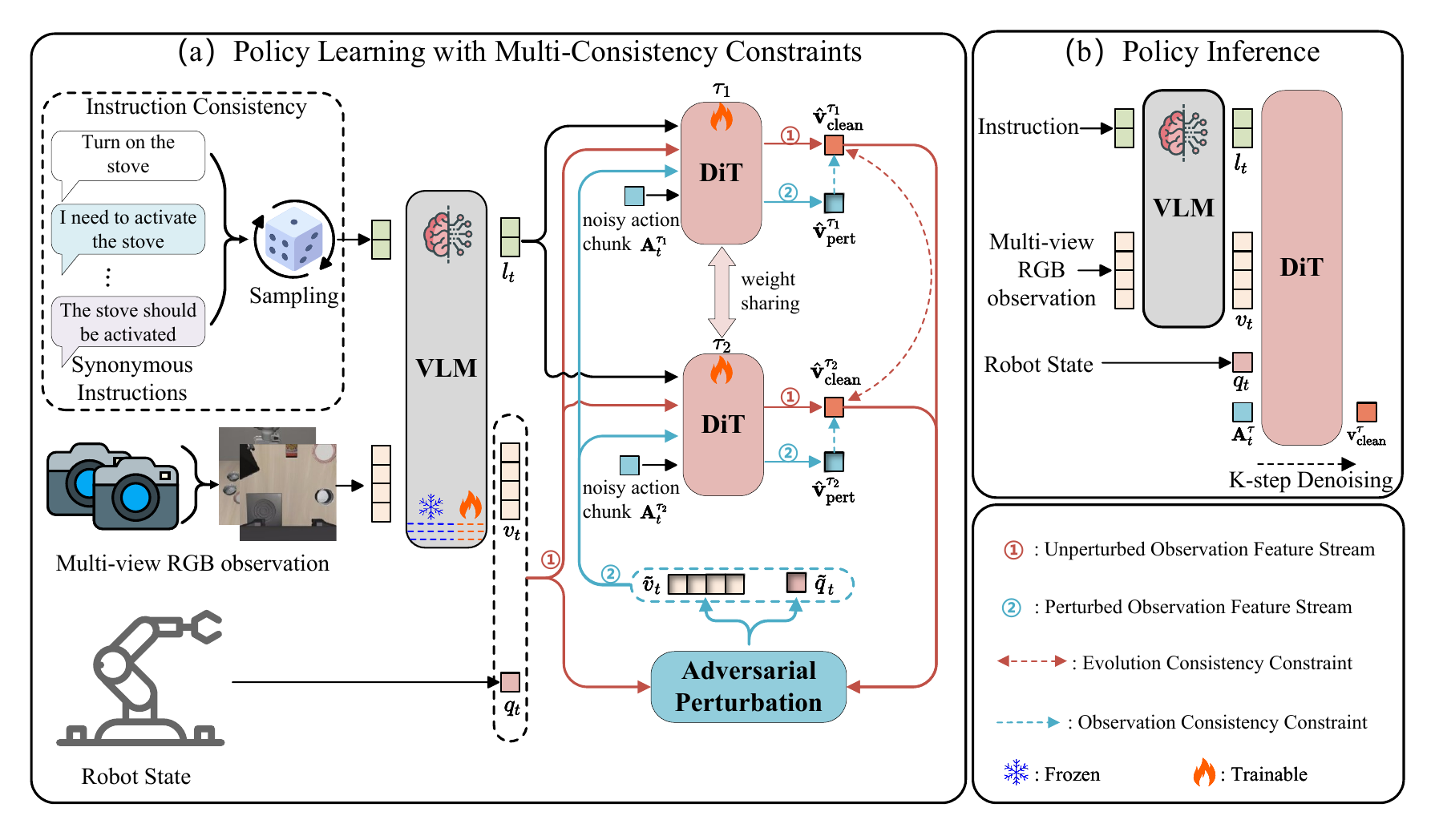}
    \vspace{-15pt}
    \caption{Overview of RoVLA. (a) RoVLA adopts a dual-system backbone with high-level semantic extraction and low-level action generation, and incorporates three consistency constraints for policy learning, including Instructional Consistency (IC), Evolutionary Consistency (EC), and Observational Consistency (OC). (b) During inference, the policy follows the standard dual-system backbone and performs $K$ denoising steps for action generation.}
    \label{fig:framework}
\end{figure*}

\section{Methodology}
\subsection{Problem formulation}
At timestep $t$, the VLA policy receives a multi-view RGB observation $\vect{I}_t\in\R^{M\times H\times W\times 3}$, a natural-language instruction $T$, and the robot state $\vect{q}_t\in\R^{d_s}$. The goal is to predict a future action chunk
\begin{equation}
\vect{A}_t=[\vect{a}_t,\vect{a}_{t+1},\ldots,\vect{a}_{t+L-1}]\in \R^{L\times d_a},
\end{equation}
where $L$ is the chunk horizon and $d_a$ is the action dimension.

We model action generation with conditional flow matching, a deterministic generative approach that enables fast sampling by learning a velocity field whose integral curves map Gaussian noise to the expert action distribution. Given an expert action chunk $\vect{A}^{\mathrm{gt}}_t$, we define a linear probability path from Gaussian noise to the target action:
\begin{equation}
\label{eq:path}
\vect{A}^{\tau}_t=\tau\vect{A}^{\mathrm{gt}}_t+(1-\tau)\bm{\epsilon},\quad \bm{\epsilon}\sim\mathcal{N}(\vect{0},\vect{I}),\quad \tau\in[0,1].
\end{equation}
The corresponding ground-truth velocity field is
\begin{equation}
\vect{v}_{\mathrm{gt}}=\vect{A}^{\mathrm{gt}}_t-\bm{\epsilon},
\end{equation}
which is invariant to $\tau$ under the linear path. A parameterized network $\vect{V}_{\theta}$ is trained to predict this conditional velocity field:
\begin{equation}
\label{eq:fm}
\mathcal{L}_{\mathrm{FM}}=\mathbb{E}_{\tau}\Big[\norm{\vect{V}_{\theta}(T,\vect{I}_t,\vect{A}^{\tau}_t,\vect{q}_t,\tau)-\vect{v}_{\mathrm{gt}}}_2^2\Big].
\end{equation}

\subsection{Dual-system VLA backbone}
RoVLA adopts a dual-system architecture inspired by recent VLA designs~\cite{nvidia2025gr00t}. A pretrained vision-language-model (VLM) serves as the semantic extractor, while a Diffusion Transformer (DiT) serves as the continuous action generator.

Given an instruction $T$ and multi-view images $\vect{I}_t$, the semantic encoder extracts aligned language and visual tokens:
\begin{equation}
[l_t, v_t] = \mathrm{VLM}_{\theta_1}(T, \vect{I}_t),
\end{equation}
where $l_t \in \mathbb{R}^{N_l \times d}$ and $v_t \in \mathbb{R}^{N_v \times d}$ denote the language and visual token sequences, respectively. We instantiate the encoder with InternVL3.5-2B~\cite{wang2025internvl3_5} and take the hidden states of its 16th decoder layer as task-conditioned semantic features, following prior findings that middle-layer VLM embeddings yield faster inference and higher downstream policy success than final-layer representations~\cite{bjorck2025gr00t}.

Following GR00T-N1.6~\cite{nvidia2025gr00t}, the action generator is implemented as a variant of DiT~\cite{peebles2023scalable} under the conditional flow-matching paradigm~\cite{lipman2022flow}. It consists of 32 Transformer layers and incorporates the denoising timestep through Adaptive Layer Normalization (AdaLN), enabling effective conditioning of the action generation process. We denote the entire action generator, including input alignment, the Transformer backbone, and the output decoding head, by $\mathrm{DiT}_{\theta_2}$.

Specifically, the robot state $\vect{q}_t$ and the noisy action chunk $\vect{A}_t^{\tau}$ are first projected by separate embedding layers into a shared latent space, forming state tokens and action tokens, respectively. The action projection is additionally conditioned on the positional encoding of the denoising timestep $\tau$, providing explicit temporal awareness for action evolution. These tokens, together with the semantic features $[l_t, v_t]$, are then processed jointly to predict the conditional velocity field:
\begin{equation}
\hat{\vect{v}}^{\tau} = \mathrm{DiT}_{\theta_2}(l_t, v_t, \vect{A}^{\tau}_t, \vect{q}_t, \tau).
\end{equation}
This backbone alone defines a standard conditional flow-matching VLA policy. As shown in Fig.~\ref{fig:framework}, We further improve its robustness during fine-tuning by introducing \textbf{multi-consistency constraints} under diverse transformations. Specifically, RoVLA incorporates three complementary constraints: \textbf{Instructional Consistency} (IC) for semantically equivalent instruction transformation, \textbf{Evolutionary Consistency} (EC) for trajectory evolution transformation, and \textbf{Observational Consistency} (OC) for observation perturbation transformation. Together, they enhance semantic invariance, action-intent stability, and robustness to input perturbations.

\subsection{Instructional consistency}
A single manipulation trajectory can often be described by multiple semantically equivalent instructions with different surface forms. Directly training on a fixed instruction for each trajectory may therefore induce a shallow coupling between specific wording patterns and action outputs, weakening generalization to paraphrased commands. To mitigate this issue, we introduce \textbf{Instructional Consistency} (IC), which promotes invariance to instruction reformulation at the data level.

For each trajectory containing only a single instruction, we construct a synonymous instructions set:
\begin{equation}
\mathcal{D}_T=\{T^{(1)},\dots,T^{(N_{\text{lang}})}\},
\end{equation}
by using Qwen3-8B~\cite{yang2025qwen3} to generate semantically equivalent paraphrases of the original instruction, where $N_{\text{lang}}$ denotes the number of semantically equivalent instructions associated with the trajectory. To enhance linguistic diversity, we generate paraphrases using multiple prompt templates and stochastic decoding with temperature and nucleus sampling. At each training iteration, one instruction is uniformly sampled from $\mathcal{D}_T$ as the language condition. This simple strategy does not introduce an additional explicit loss. Instead, it implicitly regularizes the policy to map diverse linguistic expressions to consistent task semantics. As a result, the model is encouraged to rely less on superficial wording patterns and more on the underlying task intent.

\subsection{Evolutionary consistency}
Action representations may become unstable across different stages of the flow-matching evolution process, which can in turn impair stable modeling of task intent. To address this issue, we introduce \textbf{Evolutionary Consistency} (EC) to improve the stability and continuity of velocity-field prediction over the evolution trajectory.

Specifically, within each training iteration, we sample two evolution timesteps $\tau_1,\tau_2\in[0,1]$. Under the clean input condition, the corresponding velocity predictions are computed as
\begin{align}
\hat{\vect{v}}_{\mathrm{clean}}^{\tau_1} &= \mathrm{DiT}_{\theta_2}(l_t,v_t,\vect{A}^{\tau_1}_t,\vect{q}_t,\tau_1), \\
\hat{\vect{v}}_{\mathrm{clean}}^{\tau_2} &= \mathrm{DiT}_{\theta_2}(l_t,v_t,\vect{A}^{\tau_2}_t,\vect{q}_t,\tau_2).
\end{align}
We then define the evolutionary consistency loss as
\begin{equation}
\mathcal{L}_{\mathrm{EC}}=\norm{\hat{\vect{v}}_{\mathrm{clean}}^{\tau_1}-\hat{\vect{v}}_{\mathrm{clean}}^{\tau_2}}_2^2.
\end{equation}
This pairwise formulation is sufficient to encourage smoothness along the evolution trajectory while avoiding the computational cost of multi-point supervision. Following the timestep sampling strategy of $\pi_0$~\cite{black2024pi_0}, the evolution timestep $\tau$ is sampled non-uniformly using a modified Beta distribution:
\begin{equation}
p(\tau)=\mathrm{Beta}\!\left(\frac{s-\tau}{s};\,1.5,\,1\right), \quad s=0.999.
\end{equation}
By enforcing consistency between velocity predictions at different evolution stages, EC suppresses representation fluctuations during denoising, thereby improving the stability of task-intent modeling and promoting smoother action evolution.

\subsection{Observational consistency}
% Inspired by PGD and FGSM ~\cite{madry2017towards,Goodfellow2014ExplainingAH}, To improve robustness against view shifts, background variation, lighting changes, and robot-state deviations, we introduce \textbf{observational consistency} (OC), which constructs targeted adversarial perturbations along directions where the model is most vulnerable in terms of internal consistency. 
To improve robustness to viewpoint shifts, background variations, illumination changes, and deviations in robot states, we draw inspiration from adversarial perturbation methods~\cite{madry2017towards,Goodfellow2014ExplainingAH} and introduce \textbf{Observational Consistency} (OC). OC applies targeted perturbations along the directions that most severely compromise internal consistency. Concretely, it leverages the gradients of $\mathcal{L}_{\mathrm{EC}}$ to perturb the semantic visual feature $v_t$ and the robot state $\vect{q}_t$, since perturbing semantic visual features is computationally more efficient than image-space augmentation:
\begin{align}
\tilde{v}_t &= v_t + \eta_v \frac{\nabla_{v_t}\mathcal{L}_{\mathrm{EC}}}{\norm{\nabla_{v_t}\mathcal{L}_{\mathrm{EC}}}_2},
\qquad
\eta_v = \min(\alpha,\epsilon_{\mathrm{adv}}\norm{v_t}_2),\\
\tilde{\vect{q}}_t &= \vect{q}_t + \eta_q \frac{\nabla_{\vect{q}_t}\mathcal{L}_{\mathrm{EC}}}{\norm{\nabla_{\vect{q}_t}\mathcal{L}_{\mathrm{EC}}}_2},
\qquad
\eta_q = \min(\alpha,\epsilon_{\mathrm{adv}}\norm{\vect{q}_t}_2),
\end{align}
where $\nabla_{v_t}\mathcal{L}_{\mathrm{EC}}$ and $\nabla_{\vect{q}_t}\mathcal{L}_{\mathrm{EC}}$ denote the gradients of $\mathcal{L}_{\mathrm{EC}}$ with respect to the visual feature $v_t$ and robot state$\vect{q}_t$, respectively. Here, $\alpha$ denotes the perturbation step size and $\epsilon_{\mathrm{adv}}$ controls the relative perturbation budget.

Feeding the perturbed inputs into the model yields the corresponding velocity predictions $\hat{\vect{v}}_{\mathrm{pert}}^{\tau_1}$ and $\hat{\vect{v}}_{\mathrm{pert}}^{\tau_2}$. We then define the OC loss as
\begin{equation}
\mathcal{L}_{\mathrm{OC}}
=
\frac{1}{2}\sum_{i=1}^{2}
\left\|
\hat{\vect{v}}_{\mathrm{pert}}^{\tau_i}
-
\mathrm{sg}\!\left(\hat{\vect{v}}_{\mathrm{clean}}^{\tau_i}\right)
\right\|_2^2,
\end{equation}
where $\mathrm{sg}(\cdot)$ denotes the stop-gradient operator, which prevents gradients from flowing back into the unperturbed branch so that $\hat{\vect{v}}_{\mathrm{clean}}^{\tau_i}$ acts as a fixed target and ensures the network only adjusts the perturbed predictions toward the clean ones, avoiding mutual collapse or drift during training. This objective enforces consistent velocity-field predictions before and after perturbation, thereby improving tolerance to observation shifts.

\vspace{-5pt}
\subsection{Training objective and inference}
Since $\vect{v}_{\mathrm{gt}}$ is shared across the sampled stages under Eq.~\eqref{eq:path}, the supervised losses for clean and perturbed inputs are defined as
\begin{align}
\mathcal{L}^{\mathrm{clean}} &= \frac{1}{2}\sum_{i=1}^{2}\norm{\hat{\vect{v}}_{\mathrm{clean}}^{\tau_i}-\vect{v}_{\mathrm{gt}}}_2^2,\\
\mathcal{L}^{\mathrm{pert}} &= \frac{1}{2}\sum_{i=1}^{2}\norm{\hat{\vect{v}}_{\mathrm{pert}}^{\tau_i}-\vect{v}_{\mathrm{gt}}}_2^2,
\end{align}
\begin{equation}
\mathcal{L}_{\mathrm{SFT}} = \frac{1}{2}\left(\mathcal{L}^{\mathrm{clean}}+\mathcal{L}^{\mathrm{pert}}\right).
\end{equation}
We further combine EC and OC into a consistency regularizer:
\begin{equation}
\mathcal{L}_{\mathrm{C}}=\frac{1}{2}\left(\mathcal{L}_{\mathrm{EC}}+\mathcal{L}_{\mathrm{OC}}\right).
\end{equation}
The overall training objective is
\begin{equation}
\mathcal{L}_{\mathrm{total}}=\mathcal{L}_{\mathrm{SFT}}+\lambda_j\mathcal{L}_{\mathrm{C}},
\end{equation}
where $\lambda_j$ is an adaptive weight computed from the exponential moving average of the clean supervised loss at training step $j$:
\begin{align}
\lambda_j &= \frac{1}{1+\mathcal{L}^{\mathrm{ema}}_j},\\
\mathcal{L}^{\mathrm{ema}}_j &= \gamma\mathcal{L}^{\mathrm{ema}}_{j-1}+(1-\gamma)\mathcal{L}^{\mathrm{clean}},
\end{align}
with $\mathcal{L}^{\mathrm{ema}}_0=100$ and $\gamma=0.95$. This schedule allows the model to first learn basic action prediction and then gradually place more emphasis on consistency constraints as training stabilizes, improving robustness without interfering with early-stage optimization.

At inference time, as illustrated in Fig.~\ref{fig:framework}(b), RoVLA follows the standard dual-system backbone and generates future actions through $K$ denoising steps conditioned on the current language feature $l_t$, visual feature $v_t$, and robot state $\vect{q}_t$. We first sample the initial action chunk from a standard Gaussian:
\begin{equation}
\vect{A}^{0}_t\sim\mathcal{N}(\vect{0},\vect{I}),
\end{equation}
and then iteratively update it via forward Euler integration of the learned velocity field:
\begin{equation}
\vect{A}^{\tau+\Delta\tau}_t
=
\vect{A}^{\tau}_t+\Delta\tau\cdot
\mathrm{DiT}_{\theta_2}(l_t,v_t,\vect{A}^{\tau}_t,\vect{q}_t,\tau),
\end{equation}
where $\Delta\tau=1/K$ is the integration step size. After $K$ iterations, the final denoised action trajectory is obtained.

\section{Experiments}
\subsection{Experimental Setup}

\begin{figure}[t]
    \centering
    \includegraphics[width=1.0\linewidth]{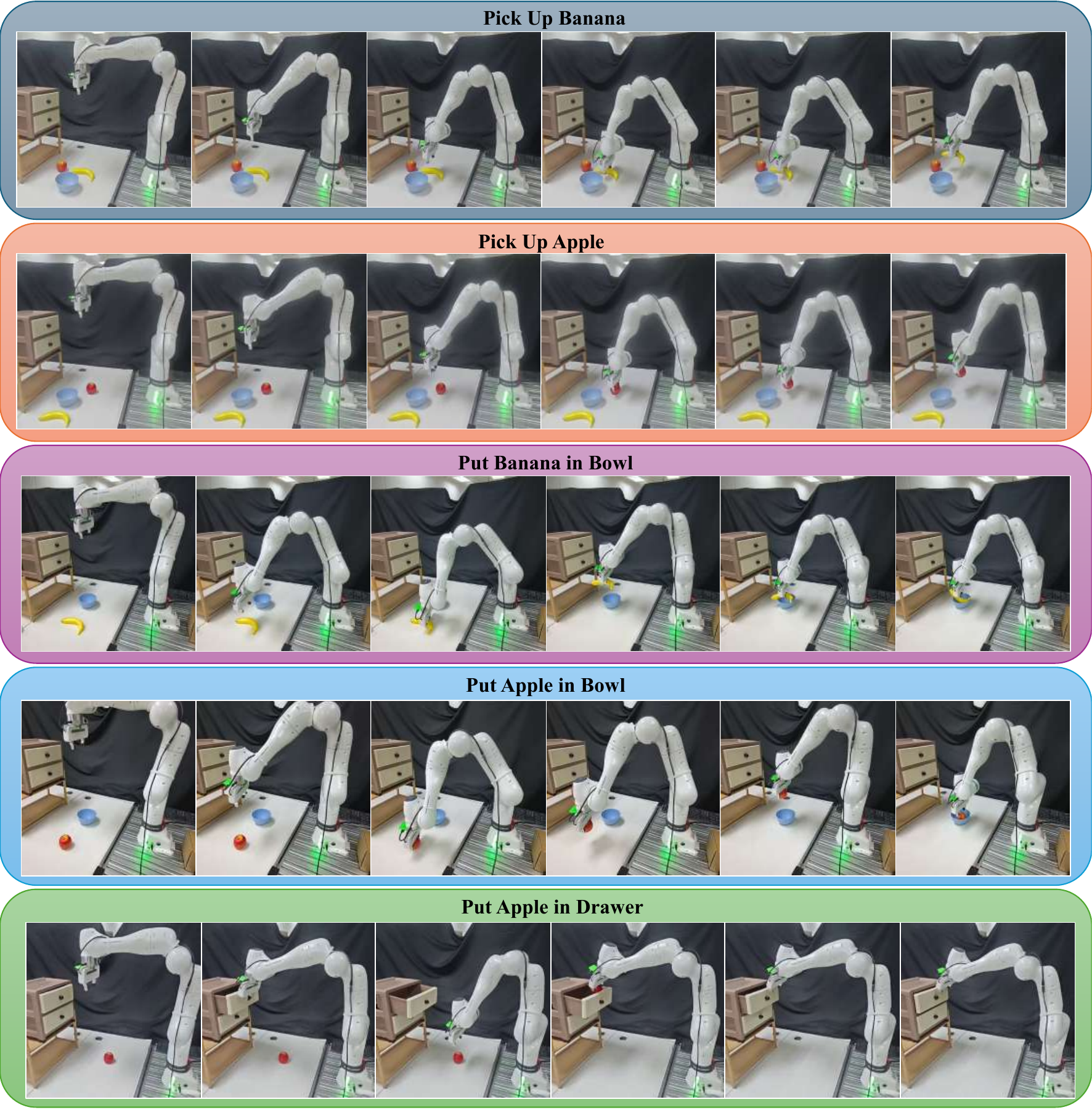}
    \caption{Visualization of the real-world evaluation tasks. We build five tabletop manipulation tasks on a Franka Research 3 robot equipped with a wrist-mounted camera to evaluate perception, instruction understanding, and manipulation performance in real-world settings.}
    \vspace{-5pt}
    \label{fig:real_world_tasks}
\end{figure}

\textbf{Benchmarks.}
We evaluate RoVLA on two challenging simulation benchmarks and a real-world robotic platform. 
\textbf{LIBERO-Plus}~\cite{fei2025libero} extends LIBERO~\cite{liu2023libero} with seven perturbation dimensions, including layout, camera, robot initialization, language, light, background, and sensor noise. We consider two evaluation settings: (1) zero-shot disturbance generalization, where models are fine-tuned on the original LIBERO demonstrations and tested directly on LIBERO-Plus, and (2) disturbance-exposed generalization, where models are fine-tuned on LIBERO-Plus yet are still required to generalize to unseen perturbation combinations at test time. 
\textbf{RoboTwin 2.0}~\cite{chen2025robotwin} contains 50 dual-arm manipulation tasks under both clean simulation environments and highly randomized environments.
To further assess transferability beyond simulation, we  construct five \textbf{real-world tabletop manipulation tasks} on a Franka Research 3 robot equipped with a wrist-mounted camera, as illustrated in Fig.~\ref{fig:real_world_tasks}. These tasks include \textit{Pick Up Banana}, \textit{Pick Up Apple}, \textit{Put Banana in Bowl}, \textit{Put Apple in Bowl}, and \textit{Put Apple in Drawer}.

\textbf{Training data and paraphrases.}
For LIBERO, we merge all four task suites, namely LIBERO-Spatial, LIBERO-Object, LIBERO-Goal, and LIBERO-Long, and use all 1,693 demonstrations for base policy learning. For LIBERO-Plus, we use its perturbation-augmented training split, which contains 15,874 successful demonstrations. Each trajectory provides third-person observations, wrist-view images, language instructions, proprioceptive states, and next-step actions. For both LIBERO and LIBERO-Plus, we use Qwen3-8B~\cite{yang2025qwen3} to generate approximately 15 distinct yet semantically equivalent paraphrases for each trajectory, which are used to support instruction consistency learning.

For RoboTwin 2.0, we collect training data in both clean and randomized simulation environments. Specifically, we collect 2,500 successful demonstrations in the clean setting (50 per task) and 25,000 successful demonstrations in the randomized setting (500 per task). The randomized environments involve background variation, object layout variation, table-height variation, and lighting variation. Unlike LIBERO-Plus, where paraphrases are constructed via post hoc rewriting, RoboTwin 2.0 natively provides 100 semantically equivalent but lexically diverse instructions for each task during data collection. we directly aggregate these native expressions as the paraphrase set for instruction consistency training.

For the real-world setting, we collect 25 successful demonstrations for each task, yielding 125 real-world trajectories in total. Consistent with LIBERO and LIBERO-Plus, we further use Qwen3-8B~\cite{yang2025qwen3} to generate approximately 15 semantically equivalent and non-duplicate paraphrases for each trajectory, enabling instruction consistency training in real-world experiments.

\begin{table*}[t]
    \centering
    \caption{Evaluation on LIBERO-Plus under two settings. The "Pretrained" column indicates whether the model was pretrained on large-scale robotic manipulation data. The best results are marked bold, and the second-best ones are underlined.}
    % \vspace{-5pt}
    \label{tab:liberoplus}
    \resizebox{\textwidth}{!}{
    \setlength{\tabcolsep}{5pt}
    \scriptsize
    \begin{tabular}{lcccccccc | c}
        \toprule
        Method & Pretrained &Camera & Robot & Language & Light & Background & Noise & Layout & Total \\
        \midrule
        \multicolumn{10}{c}{Fine-tuned on the LIBERO dataset} \\
        \midrule
        OpenVLA~\cite{kim2024openvla}  &$\surd$ &0.8 &3.5 &23.0 &8.1 &34.8 &15.2 &28.5 &15.6 \\
        OpenVLA-OFT~\cite{kim2025openvlaoft}  &$\surd$&56.4 &31.9 &79.5 &88.7 &93.3 &75.8 &\textbf{74.2} &69.6 \\
        OpenVLA-OFT\_m~\cite{kim2025openvlaoft}  &$\surd$&55.6 &21.7 &\uline{81.0} &92.7 &91.0 &78.6 &68.7 &67.9 \\
        WorldVLA~\cite{cen2025worldvla} &$\times$ & 0.1 &27.9 &41.6 &43.7 &17.1 &10.9 &38.0 &25.0 \\
        UniVLA~\cite{wang2025univla} &$\surd$ &1.8 &\textbf{46.2} &69.6 &69.0 &81.0 &21.2 &31.9 &42.9 \\
        $\pi_0$~\cite{black2024pi_0}  &$\surd$ &13.8 &6.0 &58.8 &85.0 &81.4 &79.0 &68.9 &53.6 \\
        $\pi_0$-Fast~\cite{pertsch2025fast}  &$\surd$&\textbf{65.1} &21.6 &61.0 &73.2 &73.2 &74.4 &68.8 &61.6 \\
        RIPT-VLA~\cite{tan2025ript} &$\surd$ &55.2 &31.2 &77.6 &88.4 &91.6 &73.5 &\textbf{74.2} &68.4 \\
        \midrule
        InternVL-3.5 + DiT &$\times$  &58.3 &\uline{37.2} &76.3 &\uline{95.1} &\uline{94.8} &\uline{79.1} &\uline{74.0} &\uline{71.6} \\
        \textbf{RoVLA(Ours)}&$\times$ &\uline{58.4} &36.3 &\textbf{92.9} &\textbf{95.6} &\textbf{95.0} &\textbf{80.9} &73.0 &\textbf{74.3} \\
        $\Delta$ & & +0.1 & -0.9 & +16.6 & +0.7 & +0.2 & +1.8 & -1.0 & +2.7 \\
        \midrule
        \multicolumn{10}{c}{Fine-tuned on the LIBERO-Plus dataset} \\
        \midrule
        $\pi_0$~\cite{black2024pi_0}  &$\surd$&79.6 &21.1 &72.5 &84.7 &86.2 &68.3 &69.4 &67.4\\
        $\pi_{0.5}~\cite{black2025pi_05}$  &$\surd$&70.3 &\textbf{41.7} &81.1 &\textbf{97.3} &94.6 &71.8 &\textbf{84.9} &75.7\\
        Gr00t-N1.6~\cite{nvidia2025gr00t}  &$\surd$&92.6 &\uline{33.5} &80.1 &93.6 &\uline{95.4} &93.6  &75.0 &79.4 \\
        OpenVLA-OFT~\cite{kim2025openvlaoft} &$\surd$ & 92.8 &30.3 &\uline{85.8} &94.9 &93.9 &89.3 &\uline{77.6} &\uline{79.5} \\
        \midrule
        InternVL-3.5 + DiT &$\times$  &\uline{94.2} &32.4 &64.5 &94.7 &93.0 &\uline{94.7} &74.8 &77.1 \\
        \textbf{RoVLA(Ours)}&$\times$  &\textbf{96.6} &32.0 &\textbf{91.5} &\uline{95.9} &\textbf{96.1} &\textbf{95.1} &74.1 &\textbf{82.0} \\
        $\Delta$& & +2.4 & -0.4 & +27.0 & +1.2 & +3.1 & +0.4 & -0.7 & +4.9 \\
        \bottomrule
    \end{tabular}
    }
    \label{tab:vla-method-liberoplus}
\end{table*}

\textbf{Implementation details.}
We use the first 16 decoder layers of InternVL3.5-2B~\cite{wang2025internvl3_5} as the semantic encoder, with the 16th-layer output serving as the multimodal feature interface. During fine-tuning, only the last four of these layers are unfrozen. The action generator is operating on action chunks of length 16, and is randomly initialized.
We train RoVLA with AdamW~\cite{loshchilov2017adamw}. The learning rate is linearly warmed up over the first \SI{5}{\percent} of training to a peak value of $1\times10^{-4}$ and then decayed to 0 with a cosine schedule. For OC, we set $\alpha=0.01$ and $\epsilon_{\mathrm{adv}}=0.03$. We train for 60k steps on LIBERO-Plus and the real-world tasks, and for 120k steps on RoboTwin 2.0. The entire framework is implemented in PyTorch and trained on NVIDIA H20 GPUs. During inference, action generation is performed with 8 steps of forward Euler integration. 

% More training hyperparameters are summarized in Tab.~\ref{tab:training_hyperparameters}.

% \begin{table}[htbp]
% \centering
% \caption{Training hyperparameters of RoVLA.}
% \small
% \setlength{\tabcolsep}{5pt}
% \resizebox{0.48\textwidth}{!}{
% \begin{tabular}{lcc}
% \toprule
% \textbf{Hyperparameter} & \textbf{LIBERO-Plus / Real-world} & \textbf{RoboTwin 2.0} \\
% \midrule
% Perturbation step size $\alpha$ & 0.01 & 0.01 \\
% Perturbation budget $\epsilon_{\mathrm{adv}}$ & 0.03 & 0.03 \\
% Optimizer & AdamW & AdamW \\
% Batch size & 256 & 256 \\
% Peak learning rate & $1\times10^{-4}$ & $1\times10^{-4}$ \\
% Warmup ratio & 0.05 & 0.05 \\
% Training steps & 60,000 & 120,000 \\
% AdamW betas & $(0.9,\ 0.999)$ & $(0.9,\ 0.999)$ \\
% AdamW epsilon & $1\times10^{-8}$ & $1\times10^{-8}$ \\
% Weight decay & $1\times10^{-5}$ & $1\times10^{-5}$ \\
% Gradient clipping & 1.0 & 1.0 \\
% Precision & bfloat16 & bfloat16 \\
% \bottomrule
% \end{tabular}
% }
% \label{tab:training_hyperparameters}
% \end{table}

\textbf{Metric.}
We use task success rate (\%) as the evaluation metric:
\begin{equation}
\text{Success Rate} = \frac{N_{\mathrm{succ}}}{N_{\mathrm{total}}} \times 100\%,
\end{equation}
where $N_{\mathrm{succ}}$ is the number of successful trajectories and $N_{\mathrm{total}}$ is the total number of test trajectories.

\subsection{Results on LIBERO-Plus}

\textbf{Baselines and protocol.}
We compare RoVLA with representative VLA baselines spanning diverse design paradigms, including autoregressive action modeling (e.g., OpenVLA~\cite{kim2024openvla}, OpenVLA-OFT~\cite{kim2025openvlaoft}, and OpenVLA-OFT\_m~\cite{kim2025openvlaoft} which trained uniformly on the four LIBERO task suites), flow-based action generation (e.g., $\pi_0$~\cite{black2024pi_0}, $\pi_0$-Fast~\cite{pertsch2025fast}, and $\pi_{0.5}$~\cite{black2025pi_05}), RL-based post-training (e.g., RIPT-VLA~\cite{tan2025ript}), unified token modeling (e.g., UniVLA~\cite{wang2025univla} and WorldVLA~\cite{cen2025worldvla}), and methods pretrained on large-scale cross-embodiment data (e.g., GR00T-N1.6~\cite{nvidia2025gr00t}). We further include \textbf{InternVL3.5+DiT}, our backbone-only baseline without consistency constraints, to isolate the effect of the proposed training framework. Following the official LIBERO-Plus official protocol~\cite{fei2025libero}, each method is evaluated once on all 10,300 test instances.

\textbf{Zero-shot disturbance generalization.}
When trained on LIBERO and tested directly on LIBERO-Plus, RoVLA achieves the best overall success rate of \SI{74.3}{\percent}, outperforming OpenVLA-OFT (\SI{69.6}{\percent}), RIPT-VLA (\SI{68.4}{\percent}), and $\pi_0$-Fast (\SI{61.6}{\percent}). It also improves over the backbone-only InternVL3.5+DiT by \SI{2.7}{\percent}. The largest gain appears on \textbf{Language}, where RoVLA reaches \SI{92.9}{\percent}, improving over InternVL3.5+DiT by \SI{16.6}{\percent} and over the strongest baseline by \SI{13.4}{\percent}, indicating substantially stronger invariance to paraphrased instructions. RoVLA also performs strongly on \textbf{Light}, \textbf{Background}, and \textbf{Noise}, achieving \SI{95.6}{\percent}, \SI{95.0}{\percent}, and \SI{80.9}{\percent}, respectively, which confirms the benefit of observation-level consistency training. On \textbf{Camera}, RoVLA achieves \SI{58.4}{\percent}. While this result is below $\pi_0$-Fast, the latter appears particularly strong under viewpoint variation, whereas the advantages of RoVLA are more consistently reflected in semantic robustness and broader observation perturbations. By contrast, \textbf{Layout} and \textbf{Robot} remain more challenging, as the current consistency mechanisms favor semantic and perceptual over object-layout or robot-initial invariances.

\textbf{Disturbance-exposed generalization.}
When directly fine-tuned on LIBERO-Plus, RoVLA further improves to \SI{82.0}{\percent}, achieving the best overall result. It exceeds OpenVLA-OFT (\SI{79.5}{\percent}) by 2.5 points, GR00T-N1.6 (\SI{79.4}{\percent}) by 2.6 points and InternVL3.5+DiT (\SI{77.1}{\percent}) by 4.9 points. The largest improvement again comes from \textbf{Language}, where RoVLA attains \SI{91.5}{\percent}, outperforming InternVL3.5+DiT by 27.0 points. This shows that exposure to instruction variation alone is insufficient, and that explicit consistency constraints all critical for stable instruction-conditioned control. RoVLA also achieves strong robustness on \textbf{Camera}, \textbf{Light}, \textbf{Background}, and \textbf{Noise}, reaching \SI{96.6}{\percent}, \SI{95.9}{\percent}, \SI{96.1}{\percent}, and \SI{95.1}{\percent}, respectively; among them, Camera, Background, and Noise are the best in this setting. In comparison, \textbf{Layout} and \textbf{Robot} remain more challenging. With the advantage of large-scale multi-source embodied pretraining, $\pi_{0.5}$ performs significantly better on both \textbf{Layout} and \textbf{Robot}, suggesting stronger robustness to layout perturbations and state-conditioned adaptation. RoVLA is not dominant on these two splits, but remains competitive overall even without embodied pretraining.

\textbf{Summary.}
Across both settings, RoVLA achieves the best overall performance, with the clearest gains on \textbf{Language} and observation-related perturbations. The weaker margins on \textbf{Layout} and \textbf{Robot} suggest that these splits require additional spatial and state-conditioned modeling beyond the current consistency constraints.

\begin{table}[t]
\centering
\caption{Evaluation on RoboTwin 2.0 under Clean and Randomized environments. The best results are marked in bold.}
\vspace{-5pt}
\resizebox{\columnwidth}{!}{
\begin{tabular}{lcccccccc}
\toprule
\multirow{2}{*}{Simulation Task} & \multicolumn{2}{c}{GO-1~\cite{bu2025agibot}} & \multicolumn{2}{c}{$\pi_{0.5}$~\cite{black2025pi_05}} & \multicolumn{2}{c}{InternVL-3.5+DiT} & \multicolumn{2}{c}{\textbf{RoVLA(Ours)}} \\
& Clean & Rand. & Clean & Rand. & Clean & Rand. & Clean & Rand. \\
\midrule
% \textit{Adjust Bottle} & 49.0 & 62.0 & 79.0 & 83.0 & 99.0 & 99.0 & \textbf{100.0} & \textbf{100.0} \\
% \textit{Beat Block Hammer} & 6.0 & 10.0 & \textbf{63.0} & \textbf{50.0} & 29.0 & 27.0 & 25.0 & 24.0 \\
% \textit{Blocks Ranking Rgb} & 7.0 & 3.0 & \textbf{43.0} & \textbf{35.0} & 5.0 & 10.0 & 6.0 & 19.0 \\
% \textit{Blocks Ranking Size} & 2.0 & 2.0 & \textbf{8.0} & \textbf{14.0} & 3.0 & 1.0 & 0.0 & 10.0 \\
\textit{Click Alarmclock} & 95.0 & 90.0 & \textbf{97.0} & 93.0 & 92.0 & \textbf{99.0} & 90.0 & 94.0 \\
\textit{Click Bell} & \textbf{98.0} & \textbf{95.0} & 75.0 & 76.0 & 89.0 & 85.0 & 97.0 & 94.0 \\
\textit{Dump Bin Bigbin} & 57.0 & 45.0 & 30.0 & 42.0 & \textbf{83.0} & 71.0 & 82.0 & \textbf{76.0} \\
\textit{Grab Roller} & \textbf{99.0} & \textbf{99.0} & 90.0 & 89.0 & 88.0 & 96.0 & 80.0 & 87.0 \\
% \textit{Handover Block} & 9.0 & 12.0 & \textbf{18.0} & \textbf{19.0} & 0.0 & 2.0 & 0.0 & 3.0 \\
\textit{Handover Mic} & 12.0 & 8.0 & 28.0 & 18.0 & 25.0 & 26.0 & \textbf{34.0} & \textbf{34.0} \\
% \textit{Hanging Mug} & 0.0 & 0.0 & \textbf{3.0} & \textbf{3.0} & 0.0 & 0.0 & 0.0 & 1.0 \\
% \textit{Lift Pot} & \textbf{92.0} & \textbf{92.0} & 0.0 & 0.0 & 62.0 & 41.0 & 72.0 & 59.0 \\
\textit{Move Can Pot} & 16.0 & 4.0 & 29.0 & 27.0 & 62.0 & \textbf{64.0} & \textbf{63.0} & 47.0 \\
\textit{Move Pillbottle Pad} & 9.0 & 11.0 & 33.0 & 29.0 & 39.0 & \textbf{47.0} & \textbf{42.0} & 43.0 \\
\textit{Move Playingcard Away} & 37.0 & 24.0 & 59.0 & 67.0 & \textbf{70.0} & \textbf{75.0} & 56.0 & 72.0 \\
\textit{Move Stapler Pad} & 3.0 & 4.0 & 16.0 & 18.0 & 20.0 & 14.0 & \textbf{22.0} & \textbf{19.0} \\
\textit{Open Laptop} & 65.0 & \textbf{60.0} & 19.0 & 35.0 & 49.0 & 46.0 & \textbf{75.0} & \textbf{60.0} \\
% \textit{Open Microwave} & 12.0 & 14.0 & \textbf{35.0} & \textbf{37.0} & 16.0 & 13.0 & 25.0 & 28.0 \\
% \textit{Pick Diverse Bottles} & \textbf{61.0} & \textbf{56.0} & 5.0 & 3.0 & 35.0 & 44.0 & 48.0 & 52.0 \\
% \textit{Pick Dual Bottles} & \textbf{81.0} & \textbf{74.0} & 10.0 & 6.0 & 21.0 & 24.0 & 26.0 & 41.0 \\
\textit{Place A2b Left} & 33.0 & 36.0 & 62.0 & 60.0 & 65.0 & 63.0 & \textbf{69.0} & \textbf{65.0} \\
\textit{Place A2b Right} & 31.0 & 22.0 & 62.0 & 57.0 & 52.0 & 68.0 & \textbf{64.0} & \textbf{74.0} \\
\textit{Place Bread Basket} & 47.0 & 52.0 & 48.0 & 56.0 & \textbf{62.0} & 57.0 & \textbf{62.0} & \textbf{61.0} \\
\textit{Place Bread Skillet} & 2.0 & 1.0 & 38.0 & 46.0 & 55.0 & \textbf{58.0} & \textbf{69.0} & 56.0 \\
......(50 tasks) & & & & & & & & \\
% \textit{Place Burger Fries} & \textbf{88.0} & \textbf{92.0} & 66.0 & 70.0 & 66.0 & 59.0 & 64.0 & 61.0 \\
% \textit{Place Can Basket} & \textbf{29.0} & \textbf{37.0} & 19.0 & 25.0 & 22.0 & 19.0 & 16.0 & 12.0 \\
% \textit{Place Cans Plasticbox} & \textbf{68.0} & \textbf{77.0} & 40.0 & 47.0 & 40.0 & 35.0 & 32.0 & 39.0 \\
\textit{Place Container Plate} & 73.0 & 70.0 & 71.0 & 78.0 & \textbf{95.0} & 88.0 & 92.0 & \textbf{89.0} \\
\textit{Place Dual Shoes} & 6.0 & 10.0 & 12.0 & 7.0 & 8.0 & 14.0 & \textbf{32.0} & \textbf{28.0} \\
% \textit{Place Empty Cup} & 44.0 & 39.0 & 75.0 & \textbf{86.0} & \textbf{80.0} & 76.0 & 73.0 & 79.0 \\
\textit{Place Fan} & 1.0 & 0.0 & 25.0 & 36.0 & 19.0 & 28.0 & \textbf{44.0} & \textbf{51.0} \\
\textit{Place Mouse Pad} & 15.0 & 10.0 & 21.0 & 26.0 & \textbf{27.0} & 24.0 & \textbf{27.0} & \textbf{28.0} \\
% \textit{Place Object Basket} & \textbf{48.0} & \textbf{49.0} & 43.0 & 36.0 & 10.0 & 25.0 & 10.0 & 27.0 \\
\textit{Place Object Scale} & 26.0 & 27.0 & 40.0 & 49.0 & 41.0 & 44.0 & \textbf{47.0} & \textbf{53.0} \\
\textit{Place Object Stand} & 56.0 & 63.0 & \textbf{74.0} & 65.0 & 68.0 & 68.0 & 72.0 & \textbf{76.0} \\
\textit{Place Phone Stand} & 30.0 & 37.0 & 49.0 & 53.0 & 55.0 & 57.0 & \textbf{59.0} & \textbf{66.0} \\
\textit{Place Shoe} & 15.0 & 13.0 & 57.0 & 61.0 & 72.0 & 68.0 & \textbf{80.0} & \textbf{72.0} \\
% \textit{Press Stapler} & 66.0 & 51.0 & \textbf{80.0} & \textbf{70.0} & 71.0 & 66.0 & 67.0 & 59.0 \\
% \textit{Put Bottles Dustbin} & 7.0 & 4.0 & \textbf{12.0} & \textbf{9.0} & 0.0 & 2.0 & 1.0 & 3.0 \\
% \textit{Put Object Cabinet} & \textbf{60.0} & \textbf{43.0} & 24.0 & 15.0 & 20.0 & 17.0 & 14.0 & 19.0 \\
\textit{Rotate Qrcode} & 22.0 & 9.0 & 47.0 & 56.0 & \textbf{51.0} & 56.0 & 48.0 & \textbf{64.0} \\
\textit{Scan Object} & 1.0 & 2.0 & 42.0 & 38.0 & \textbf{52.0} & 29.0 & \textbf{52.0} & \textbf{44.0} \\
\textit{Shake Bottle Horizontally} & \textbf{97.0} & 92.0 & 96.0 & \textbf{100.0} & 90.0 & 89.0 & 89.0 & 94.0 \\
% \textit{Shake Bottle} & \textbf{97.0} & 93.0 & 91.0 & \textbf{100.0} & 83.0 & 86.0 & 89.0 & 88.0 \\
% \textit{Stack Blocks Three} & 1.0 & 1.0 & \textbf{15.0} & \textbf{16.0} & 0.0 & 8.0 & 1.0 & 4.0 \\
% \textit{Stack Blocks Two} & 12.0 & 22.0 & \textbf{48.0} & \textbf{56.0} & 10.0 & 29.0 & 16.0 & 45.0 \\
% \textit{Stack Bowls Three} & 4.0 & 7.0 & \textbf{33.0} & \textbf{35.0} & 16.0 & 22.0 & 20.0 & 13.0 \\
% \textit{Stack Bowls Two} & 51.0 & 45.0 & \textbf{78.0} & 66.0 & 62.0 & 62.0 & 72.0 & \textbf{72.0} \\
% \textit{Stamp Seal} & 19.0 & 13.0 & 36.0 & 23.0 & 41.0 & 38.0 & \textbf{49.0} & \textbf{51.0} \\
% \textit{Turn Switch} & 34.0 & 30.0 & 5.0 & 6.0 & 26.0 & 29.0 & \textbf{39.0} & \textbf{46.0} \\
\midrule
Total & 37.8 & 36.2 & 43.0 & 43.8 & 44.9 & 45.4 & \textbf{48.2} & \textbf{50.0} \\
\bottomrule
\end{tabular}
}
\label{tab:robotwin2}
\vspace{-10pt}
\end{table}

\subsection{Results on RoboTwin 2.0}

\textbf{Baselines and protocol.}
We further compare RoVLA with representative baselines on RoboTwin 2.0, including GO-1~\cite{bu2025agibot}, which learns latent action representations from large-scale human operation videos, $\pi_{0.5}$~\cite{black2025pi_05}, which improves transfer via heterogeneous multi-task training, and InternVL3.5+DiT, our backbone-only baseline without consistency constraints. Following the official RoboTwin 2.0 protocol~\cite{chen2025robotwin}, each task is evaluated for 100 trials under both Clean and Randomized environments, and the average success rate is reported.

\textbf{Results Analysis.}
As shown in Table~\ref{tab:robotwin2}, RoVLA achieves the best overall performance in both settings, reaching \SI{48.2}{\percent} in the Clean environment and \SI{50.0}{\percent} in the Randomized environment. Compared with the same InternVL3.5+DiT backbone, this achieves gains of 3.3 and 4.6 points, respectively, and compared with $\pi_{0.5}$, the gains are 5.2 and 6.2 points. These results indicate that the proposed multi-consistency training provides reliable benefits across different tasks and remains robust to environmental variations.

Notably, RoVLA performs even better in the Randomized environment than in the Clean one. This trend suggests that the proposed consistency constraints reduce over-reliance on specific scene configurations and improve policy stability under combined shifts in appearance, pose, and scene layout.

At a high level, the gains of RoVLA are most evident on tasks requiring robust target grounding, spatial relation understanding, and adaptation to scene variation (e.g., \textit{Open Laptop}, \textit{Place A2b Left}, \textit{Scan Object} etc.), while tasks dominated by fine contact dynamics or complex bimanual coordination remain challenging (e.g., \textit{Click Alarmclock}, \textit{Move Playingcard Away}, \textit{Grab Roller} etc.). Overall, the RoboTwin 2.0 results corroborate the findings on LIBERO-Plus: enforcing consistency in instruction semantics, observation features, and action evolution improves both robustness and generalization in complex manipulation settings.

\begin{table}[t]
\centering
\caption{Evaluation on real-world tasks. The best results are marked in bold.}
\vspace{-5pt}
\label{tab:real_world_results}
\resizebox{\columnwidth}{!}{
\begin{tabular}{lccc}
\toprule
\textbf{Task} & \textbf{GR00T-N1.6} & \textbf{InternVL3.5+DiT} & \textbf{RoVLA (Ours)} \\
\midrule
\textit{Pick Up Banana} & \textbf{80} & 60 & \textbf{80} \\
\textit{Pick Up Apple} & 50 & 60 & \textbf{70} \\
\textit{Put Banana in Bowl} & 70 & 50 & \textbf{80} \\
\textit{Put Apple in Bowl} & 40 & 20 & \textbf{50} \\
\textit{Put Apple in Drawer} & 10 & 0 & \textbf{20} \\
\midrule
\textbf{Total} & 50 & 38 & \textbf{60} \\
\bottomrule
\end{tabular}
}
\end{table}

\vspace{-5pt}
\subsection{Results in the Real World}

\textbf{Baselines and protocol.}
To enable a fair comparison, we compare RoVLA with InternVL3.5+DiT, our backbone-only baseline without consistency constraints, and GR00T-N1.6~\cite{nvidia2025gr00t}, a representative VLA baseline with a comparable parameter scale that is pretrained on large-scale cross-embodiment data. Evaluation is conducted on five real-world tabletop manipulation tasks, with 10 trials per task, and average success rate is reported.

\textbf{Results Analysis.}
As shown in Table~\ref{tab:real_world_results}, RoVLA achieves the best overall success rate of \SI{60}{\percent}, outperforming GR00T-N1.6 (\SI{50}{\percent}) and InternVL3.5+DiT (\SI{38}{\percent}). RoVLA attains the best performance on four of the five tasks and ties with GR00T-N1.6 on \textit{pick up banana}. Compared with InternVL3.5+DiT, RoVLA improves performance on all tasks, indicating that the gain comes from the proposed consistency training rather than the backbone alone. Tasks involving apples are generally more challenging, suggesting higher demands on grasp precision and contact stability. In particular, \textit{put apple in drawer} is the most difficult task for all methods, likely because it requires both long-horizon manipulation and accurate grasping of a small rounded drawer handle. Even in this case, RoVLA still achieves the best result, further supporting its robustness in complex real-world settings.

\begin{table}[t]
    \centering
    \caption{Ablation on LIBERO-Plus. The best results are marked in bold.}
    % \vspace{-5pt}
    \label{tab:ablation}
    % \scriptsize
    \resizebox{\columnwidth}{!}{%
    \begin{tabular}{ccccccccccc}
        \toprule
        IC & EC & OC & Camera & Robot & Language & Light & Background & Noise & Layout & Total \\
        \midrule
        \xmark & \xmark & \xmark & 94.2 & 32.4 & 64.5 & 94.7 & 93.0 & 94.7 & 74.8 & 77.1 \\
        \xmark & \cmark & \xmark & 94.6 & 32.5 & 68.4 & 94.0 & 94.5 & 94.2 & \textbf{77.0} & 78.2 \\
        \cmark & \cmark & \xmark & 93.7 & 30.6 & 89.5 & 94.7 & 93.5 & 94.6 & 73.8 & 80.5 \\
        \xmark & \cmark & \cmark & 95.3 & \textbf{35.6} & 69.6 & 94.5 & 95.4 & \textbf{95.6} & 74.5 & 79.0 \\
        \cmark & \cmark & \cmark & \textbf{96.6} & 32.0 & \textbf{91.5} & \textbf{95.9} & \textbf{96.1} & 95.1 & 74.1 & \textbf{82.0} \\
        \bottomrule
    \end{tabular}
    }
    \vspace{-15pt}
\end{table}

\subsection{Ablation study}
We conduct a module-wise ablation on LIBERO-Plus using the InternVL3.5+DiT backbone. Results are shown in Table~\ref{tab:ablation}. Starting from the backbone without any consistency term, EC alone improves the overall success rate from \SI{77.1}{\percent} to \SI{78.2}{\percent} and yields the strongest gain on Layout, suggesting that stabilizing the denoising dynamics helps under spatial reconfiguration. Its limited gains on Robot and Noise further suggest that action-evolution consistency mainly mitigates intent drift during generation, but cannot directly address failures caused by embodiment mismatch or observation corruption. Adding IC on top of EC dramatically boosts the Language split from \SI{68.4}{\percent} to \SI{89.5}{\percent} and lifts the total success rate to \SI{80.5}{\percent}, confirming that paraphrase-based instruction consistency is the dominant factor for robust language grounding. The slight drops on several non-language splits indicate that stronger invariance to instruction reformulation does not automatically improve robustness to visual or robot-state perturbations, and may even reduce sensitivity to task-critical observation cues when optimized in isolation.

Combining EC and OC improves Robot and Noise most noticeably, indicating that the targeted perturbation strategy indeed strengthens robustness to state errors and sensor corruption. Its relatively limited gains on Language and Layout suggest that observation-side hardening mainly improves local perturbation robustness, but does not directly resolve semantic ambiguity or higher-level spatial compositionality. The full model with IC+EC+OC achieves the best overall result of \SI{82.0}{\percent}. Importantly, the best configuration is not obtained by any single module alone. Instead, the three modules interact synergistically, with IC stabilizing task semantics, EC stabilizing action intent, and OC hardening the policy against fragile observation directions. As a result, although individual modules may introduce slight trade-offs on specific splits, their combination yields the most balanced robustness across diverse perturbation settings.

\subsection{Qualitative analysis}
\begin{figure}[t]
    \centering
    \includegraphics[width=\linewidth]{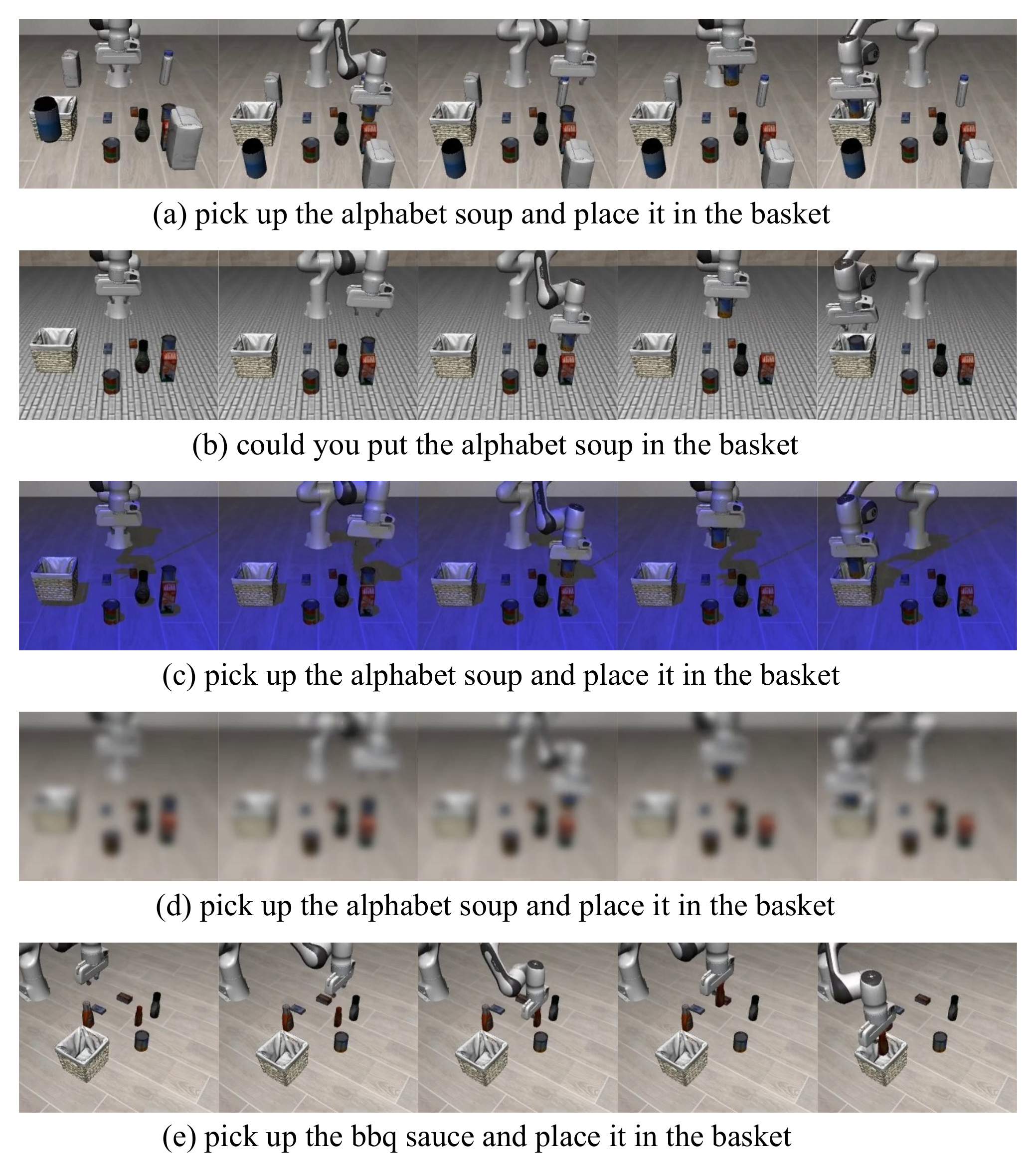}%
    \vspace{-5pt}
    \caption{Qualitative examples on LIBERO-Plus. Under various perturbation conditions, including (a) increased object count, (b) variations in language instruction expression, (c) illumination changes, (d) injected camera blur, and (e) viewpoint changes, RoVLA still maintains relatively stable action decisions.}
    \vspace{-15pt}
    \Description{Representative qualitative results of RoVLA under clutter, paraphrased instructions, lighting changes, image noise, and viewpoint shifts.}
    \label{fig:qualitative-liberoplus}
\end{figure}

Figure~\ref{fig:qualitative-liberoplus} shows representative rollout snapshots on LIBERO-Plus. Under diverse perturbation conditions, including additional clutter, paraphrased instructions, illumination changes, image blur, and camera viewpoint shifts, RoVLA maintains stable action decisions and largely preserves the intended manipulation trajectory. These qualitative results are consistent with the quantitative improvements on the Language, Camera, and observation-shift splits, further suggesting that multi-consistency enhances not only overall task success but also the stability of policy execution under distribution shifts.

\vspace{-5pt}
\section{Conclusion and Limitation}
In this paper, we proposed RoVLA, a robust vision-language-action framework with multi-consistency constraints. Built on a dual-system InternVL3.5+DiT backbone, RoVLA improves the robustness of VLA policies by jointly introducing Instructional Consistency, Evolutionary Consistency, and Observational Consistency during training. Extensive experiments on LIBERO-Plus, RoboTwin 2.0, and real-world manipulation tasks demonstrate that RoVLA consistently outperforms strong baselines, with particularly notable gains under language variation and observation perturbation. These results show that explicitly enforcing consistency across instruction semantics, action evolution, and observation perturbation provides an effective way to improve the robustness and generalization of VLA policies.

Nevertheless, RoVLA still yields limited improvements in scenarios involving fine-grained contact dynamics and complex bimanual coordination. Future work will extend the proposed multi-consistency constrained framework toward more contact-aware policies and better bimanual coordination, for example by incorporating stronger geometric and dynamics constraints into action evolution and observation modeling.

% \begin{acks}
% \end{acks}

% \bibliographystyle{ACM-Reference-Format}
\bibliographystyle{ieeetr}
\bibliography{bibfile}

% % %%
% % %% If your work has an appendix, this is the place to put it.
\appendix

\section{More Implementation details of Instructional Consistency}

We provide more implementation details for the instruction rewriting process employed by "Instruction Consistency" (IC). IC does not introduce an additional explicit loss. Instead, it expands each single-instruction trajectory with a set of diverse instruction reformulations and uniformly samples one instruction during training.

\textbf{Paraphrase generation model.}
We use Qwen3-8B~\cite{yang2025qwen3} as the instruction rewriter to construct diverse language conditions for each trajectory. We disable the model's thinking mode during generation and use stochastic decoding to increase linguistic diversity.

\textbf{Decoding setup.}
For each prompt, candidate reformulations are generated with sampling enabled (\texttt{do\_sample=True}) using temperature \(\tau_{\text{temp}}=0.7\) and nucleus sampling with \(\texttt{top\_p}=0.9\). The maximum generation length is set to \texttt{max\_new\_tokens}=512.

\textbf{Prompt templates.}
Given an original instruction \(T\), we query Qwen3-8B with seven prompt templates designed to induce surface-form diversity while preserving task intent as much as possible. For each template, we sample up to five candidate outputs. The exact prompt templates are listed below, where \texttt{\{T\}} denotes the original instruction.

\begin{enumerate}
    \item \textbf{User-intent style}
    \begin{quote}\small\ttfamily
    Rewrite the robotic command "\{T\}" as a natural expression of user intent, without using imperative mood or "please". Output only the rewritten sentence.
    \end{quote}

    \item \textbf{Functional-goal style}
    \begin{quote}\small\ttfamily
    Rephrase "\{T\}" as a functional goal that describes what should be achieved, not how to do it. Output only the rewritten sentence.
    \end{quote}

    \item \textbf{Functional-reference style}
    \begin{quote}\small\ttfamily
    Rewrite "\{T\}" by referring to objects based on their function or typical use instead of their names, do not mention the object name. Output only the rewritten sentence.
    \end{quote}

    \item \textbf{Polite-request style}
    \begin{quote}\small\ttfamily
    Rewrite "\{T\}" as a polite and courteous request a human might naturally say. Output only the rewritten sentence.
    \end{quote}

    \item \textbf{Concise-command style}
    \begin{quote}\small\ttfamily
    Make "\{T\}" as concise as possible while keeping all essential actions and objects. Output only the rewritten sentence.
    \end{quote}

    \item \textbf{Teaching style}
    \begin{quote}\small\ttfamily
    Explain the task "\{T\}" clearly as if teaching a new robot learner, focusing on purpose and clarity. Output only the rewritten sentence.
    \end{quote}

    \item \textbf{Abstract-goal style}
    \begin{quote}\small\ttfamily
    Summarize the core objective of "\{T\}" at a high level, ignoring low-level details. Output only the rewritten sentence.
    \end{quote}
\end{enumerate}

\textbf{Candidate construction and post-processing.}
Let \(g_m^{(k)}(T)\) denote the \(k\)-th sampled output under the \(m\)-th prompt template, where \(m\in\{1,\dots,7\}\) and \(k\in\{1,\dots,5\}\). We first collect the original instruction together with the generated candidates:
\begin{equation}
\widetilde{\mathcal{D}}_T=\{T\}\cup \bigcup_{m=1}^{7}\bigcup_{k=1}^{5}\left\{g_m^{(k)}(T)\right\}.
\end{equation}
Because generation is stochastic and subsequent filtering is applied, the number of valid reformulations retained from each prompt may be smaller than five. The final instruction set is obtained by lightweight filtering and deduplication:
\begin{equation}
\mathcal{D}_T=\mathrm{Dedup}\!\left(\mathrm{Filter}\!\left(\widetilde{\mathcal{D}}_T\right)\right).
\end{equation}

Concretely, we apply the following post-processing rules:
\begin{itemize}
    \item The phrase ``The user wants'' is normalized to ``I want'';
    \item Empty outputs and outputs shorter than 8 characters are discarded;
    \item Generations beginning with refusal or error-like patterns (e.g., ``sorry'', ``I cannot'', ``I don't'', ``unable'', ``error'') are discarded;
    \item Duplicate reformulations are removed.
\end{itemize}

\textbf{Instruction rewriting examples.}
A representative example of the instruction rewriting process used in IC is shown below. Starting from the original instruction
\texttt{``turn on the stove and put the moka pot on it''},
representative reformulations include:
\begin{enumerate}
    \item \textbf{User intent:} \texttt{I want the stove to be turned on and the moka pot placed on it}.
    \item \textbf{Functional goal:} \texttt{Prepare a beverage using a moka pot by heating it on the stove}.
    \item \textbf{Concise command:} \texttt{Heat the moka pot using the stove}.
    \item \textbf{Object-function reference:} \texttt{Activate the heat source and place the coffee brewing device on it}.
    \item \textbf{Polite request:} \texttt{Could you please turn on the stove and place the moka pot on it}.
    \item \textbf{Teaching style:} \texttt{Activate the stove and position the moka pot on the stove to begin the brewing process}.
    \item \textbf{Abstract goal:} \texttt{The core objective is to prepare coffee using a moka pot}.
\end{enumerate}

\textbf{Dataset integration and training usage.}
For each trajectory, the instruction reformulation set is constructed offline during data preprocessing and incorporated into the training data beforehand. During training, one instruction is uniformly sampled from \(\mathcal{D}_T\) at each iteration and used as the language condition. In this way, IC does not introduce an additional explicit loss, but instead implicitly regularizes the policy to map diverse linguistic expressions to consistent task semantics.

\section{Detailed results on RoboTwin 2.0}

Due to space constraints, the main paper reports only a subset of representative RoboTwin 2.0~\cite{chen2025robotwin} tasks together with the overall average success rate. For completeness, Table~\ref{tab:robotwin2-full} presents the full results on all 50 tasks under both Clean and Randomized environments.

Overall, the detailed breakdown is consistent with the conclusions in the main paper. RoVLA achieves the best average performance in both settings, reaching \SI{48.2}{\percent} in the Clean environment and \SI{50.0}{\percent} in the Randomized environment. Compared with the same InternVL-3.5+DiT backbone~\cite{wang2025internvl3_5, peebles2023scalable, nvidia2025gr00t}, the gains are 3.3 and 4.6 points, respectively, and the margin is larger over $\pi_{0.5}$, with improvements of 5.2 and 6.2 points. The larger advantage in the Randomized environment supports that the proposed multi-consistency training improves policy robustness under simultaneous variation in object pose, scene layout, and visual appearance.

At the task level, the benefits of RoVLA are more evident on tasks that require robust target grounding, spatial relation understanding, and adaptation to changed scene configurations, such as \textit{Open Laptop}, \textit{Place A2b Left}, \textit{Place A2b Right}, \textit{Place Fan}, \textit{Place Phone Stand}, \textit{Rotate Qrcode}, \textit{Stamp Seal}, and \textit{Turn Switch}. In several cases, RoVLA also maintains competitive or even improved performance under randomization relative to the Clean setting, which is consistent with reduced overfitting to canonical training configurations. Meanwhile, all methods, including RoVLA, remain challenged on tasks dominated by precise contact dynamics, ranking/stacking, or complex bimanual coordination, such as \textit{Beat Block Hammer}, \textit{Blocks Ranking Size}, \textit{Handover Block}, and \textit{Stack Blocks Three}. 

\begin{table*}[t]
\centering
\caption{Detailed results for all RoboTwin 2.0 tasks under Clean and Randomized environments. The best results are marked in bold.}
\setlength{\tabcolsep}{13pt}
\begin{tabular}{lcccccccc}
\toprule
\multirow{2}{*}{Simulation Task} & \multicolumn{2}{c}{GO-1~\cite{bu2025agibot}} & \multicolumn{2}{c}{$\pi_{0.5}$~\cite{black2025pi_05}} & \multicolumn{2}{c}{InternVL-3.5+DiT} & \multicolumn{2}{c}{RoVLA (Ours)} \\
& Clean & Rand. & Clean & Rand. & Clean & Rand. & Clean & Rand. \\
\midrule
\textit{Adjust Bottle} & 49.0 & 62.0 & 79.0 & 83.0 & 99.0 & 99.0 & \textbf{100.0} & \textbf{100.0} \\
\textit{Beat Block Hammer} & 6.0 & 10.0 & \textbf{63.0} & \textbf{50.0} & 29.0 & 27.0 & 25.0 & 24.0 \\
\textit{Blocks Ranking Rgb} & 7.0 & 3.0 & \textbf{43.0} & \textbf{35.0} & 5.0 & 10.0 & 6.0 & 19.0 \\
\textit{Blocks Ranking Size} & 2.0 & 2.0 & \textbf{8.0} & \textbf{14.0} & 3.0 & 1.0 & 0.0 & 10.0 \\
\textit{Click Alarmclock} & 95.0 & 90.0 & \textbf{97.0} & 93.0 & 92.0 & \textbf{99.0} & 90.0 & 94.0 \\
\textit{Click Bell} & \textbf{98.0} & \textbf{95.0} & 75.0 & 76.0 & 89.0 & 85.0 & 97.0 & 94.0 \\
\textit{Dump Bin Bigbin} & 57.0 & 45.0 & 30.0 & 42.0 & \textbf{83.0} & 71.0 & 82.0 & \textbf{76.0} \\
\textit{Grab Roller} & \textbf{99.0} & \textbf{99.0} & 90.0 & 89.0 & 88.0 & 96.0 & 80.0 & 87.0 \\
\textit{Handover Block} & 9.0 & 12.0 & \textbf{18.0} & \textbf{19.0} & 0.0 & 2.0 & 0.0 & 3.0 \\
\textit{Handover Mic} & 12.0 & 8.0 & 28.0 & 18.0 & 25.0 & 26.0 & \textbf{34.0} & \textbf{34.0} \\
\textit{Hanging Mug} & 0.0 & 0.0 & \textbf{3.0} & \textbf{3.0} & 0.0 & 0.0 & 0.0 & 1.0 \\
\textit{Lift Pot} & \textbf{92.0} & \textbf{92.0} & 0.0 & 0.0 & 62.0 & 41.0 & 72.0 & 59.0 \\
\textit{Move Can Pot} & 16.0 & 4.0 & 29.0 & 27.0 & 62.0 & \textbf{64.0} & \textbf{63.0} & 47.0 \\
\textit{Move Pillbottle Pad} & 9.0 & 11.0 & 33.0 & 29.0 & 39.0 & \textbf{47.0} & \textbf{42.0} & 43.0 \\
\textit{Move Playingcard Away} & 37.0 & 24.0 & 59.0 & 67.0 & \textbf{70.0} & \textbf{75.0} & 56.0 & 72.0 \\
\textit{Move Stapler Pad} & 3.0 & 4.0 & 16.0 & 18.0 & 20.0 & 14.0 & \textbf{22.0} & \textbf{19.0} \\
\textit{Open Laptop} & 65.0 & \textbf{60.0} & 19.0 & 35.0 & 49.0 & 46.0 & \textbf{75.0} & \textbf{60.0} \\
\textit{Open Microwave} & 12.0 & 14.0 & \textbf{35.0} & \textbf{37.0} & 16.0 & 13.0 & 25.0 & 28.0 \\
\textit{Pick Diverse Bottles} & \textbf{61.0} & \textbf{56.0} & 5.0 & 3.0 & 35.0 & 44.0 & 48.0 & 52.0 \\
\textit{Pick Dual Bottles} & \textbf{81.0} & \textbf{74.0} & 10.0 & 6.0 & 21.0 & 24.0 & 26.0 & 41.0 \\
\textit{Place A2b Left} & 33.0 & 36.0 & 62.0 & 60.0 & 65.0 & 63.0 & \textbf{69.0} & \textbf{65.0} \\
\textit{Place A2b Right} & 31.0 & 22.0 & 62.0 & 57.0 & 52.0 & 68.0 & \textbf{64.0} & \textbf{74.0} \\
\textit{Place Bread Basket} & 47.0 & 52.0 & 48.0 & 56.0 & \textbf{62.0} & 57.0 & \textbf{62.0} & \textbf{61.0} \\
\textit{Place Bread Skillet} & 2.0 & 1.0 & 38.0 & 46.0 & 55.0 & \textbf{58.0} & \textbf{69.0} & 56.0 \\
\textit{Place Burger Fries} & \textbf{88.0} & \textbf{92.0} & 66.0 & 70.0 & 66.0 & 59.0 & 64.0 & 61.0 \\
\textit{Place Can Basket} & \textbf{29.0} & \textbf{37.0} & 19.0 & 25.0 & 22.0 & 19.0 & 16.0 & 12.0 \\
\textit{Place Cans Plasticbox} & \textbf{68.0} & \textbf{77.0} & 40.0 & 47.0 & 40.0 & 35.0 & 32.0 & 39.0 \\
\textit{Place Container Plate} & 73.0 & 70.0 & 71.0 & 78.0 & \textbf{95.0} & 88.0 & 92.0 & \textbf{89.0} \\
\textit{Place Dual Shoes} & 6.0 & 10.0 & 12.0 & 7.0 & 8.0 & 14.0 & \textbf{32.0} & \textbf{28.0} \\
\textit{Place Empty Cup} & 44.0 & 39.0 & 75.0 & \textbf{86.0} & \textbf{80.0} & 76.0 & 73.0 & 79.0 \\
\textit{Place Fan} & 1.0 & 0.0 & 25.0 & 36.0 & 19.0 & 28.0 & \textbf{44.0} & \textbf{51.0} \\
\textit{Place Mouse Pad} & 15.0 & 10.0 & 21.0 & 26.0 & \textbf{27.0} & 24.0 & \textbf{27.0} & \textbf{28.0} \\
\textit{Place Object Basket} & \textbf{48.0} & \textbf{49.0} & 43.0 & 36.0 & 10.0 & 25.0 & 10.0 & 27.0 \\
\textit{Place Object Scale} & 26.0 & 27.0 & 40.0 & 49.0 & 41.0 & 44.0 & \textbf{47.0} & \textbf{53.0} \\
\textit{Place Object Stand} & 56.0 & 63.0 & \textbf{74.0} & 65.0 & 68.0 & 68.0 & 72.0 & \textbf{76.0} \\
\textit{Place Phone Stand} & 30.0 & 37.0 & 49.0 & 53.0 & 55.0 & 57.0 & \textbf{59.0} & \textbf{66.0} \\
\textit{Place Shoe} & 15.0 & 13.0 & 57.0 & 61.0 & 72.0 & 68.0 & \textbf{80.0} & \textbf{72.0} \\
\textit{Press Stapler} & 66.0 & 51.0 & \textbf{80.0} & \textbf{70.0} & 71.0 & 66.0 & 67.0 & 59.0 \\
\textit{Put Bottles Dustbin} & 7.0 & 4.0 & \textbf{12.0} & \textbf{9.0} & 0.0 & 2.0 & 1.0 & 3.0 \\
\textit{Put Object Cabinet} & \textbf{60.0} & \textbf{43.0} & 24.0 & 15.0 & 20.0 & 17.0 & 14.0 & 19.0 \\
\textit{Rotate Qrcode} & 22.0 & 9.0 & 47.0 & 56.0 & \textbf{51.0} & 56.0 & 48.0 & \textbf{64.0} \\
\textit{Scan Object} & 1.0 & 2.0 & 42.0 & 38.0 & \textbf{52.0} & 29.0 & \textbf{52.0} & \textbf{44.0} \\
\textit{Shake Bottle Horizontally} & \textbf{97.0} & 92.0 & 96.0 & \textbf{100.0} & 90.0 & 89.0 & 89.0 & 94.0 \\
\textit{Shake Bottle} & \textbf{97.0} & 93.0 & 91.0 & \textbf{100.0} & 83.0 & 86.0 & 89.0 & 88.0 \\
\textit{Stack Blocks Three} & 1.0 & 1.0 & \textbf{15.0} & \textbf{16.0} & 0.0 & 8.0 & 1.0 & 4.0 \\
\textit{Stack Blocks Two} & 12.0 & 22.0 & \textbf{48.0} & \textbf{56.0} & 10.0 & 29.0 & 16.0 & 45.0 \\
\textit{Stack Bowls Three} & 4.0 & 7.0 & \textbf{33.0} & \textbf{35.0} & 16.0 & 22.0 & 20.0 & 13.0 \\
\textit{Stack Bowls Two} & 51.0 & 45.0 & \textbf{78.0} & 66.0 & 62.0 & 62.0 & 72.0 & \textbf{72.0} \\
\textit{Stamp Seal} & 19.0 & 13.0 & 36.0 & 23.0 & 41.0 & 38.0 & \textbf{49.0} & \textbf{51.0} \\
\textit{Turn Switch} & 34.0 & 30.0 & 5.0 & 6.0 & 26.0 & 29.0 & \textbf{39.0} & \textbf{46.0} \\
\midrule
Total & 37.8 & 36.2 & 43.0 & 43.8 & 44.9 & 45.4 & \textbf{48.2} & \textbf{50.0} \\
\bottomrule
\end{tabular}
\label{tab:robotwin2-full}
\end{table*}

\section{More qualitative analysis}

We provide additional rollout visualizations on RoboTwin 2.0 and the real-world evaluation tasks.

\subsection{Qualitative analysis on RoboTwin 2.0}

\begin{figure*}[t]
    \centering
    \includegraphics[width=0.95\linewidth]{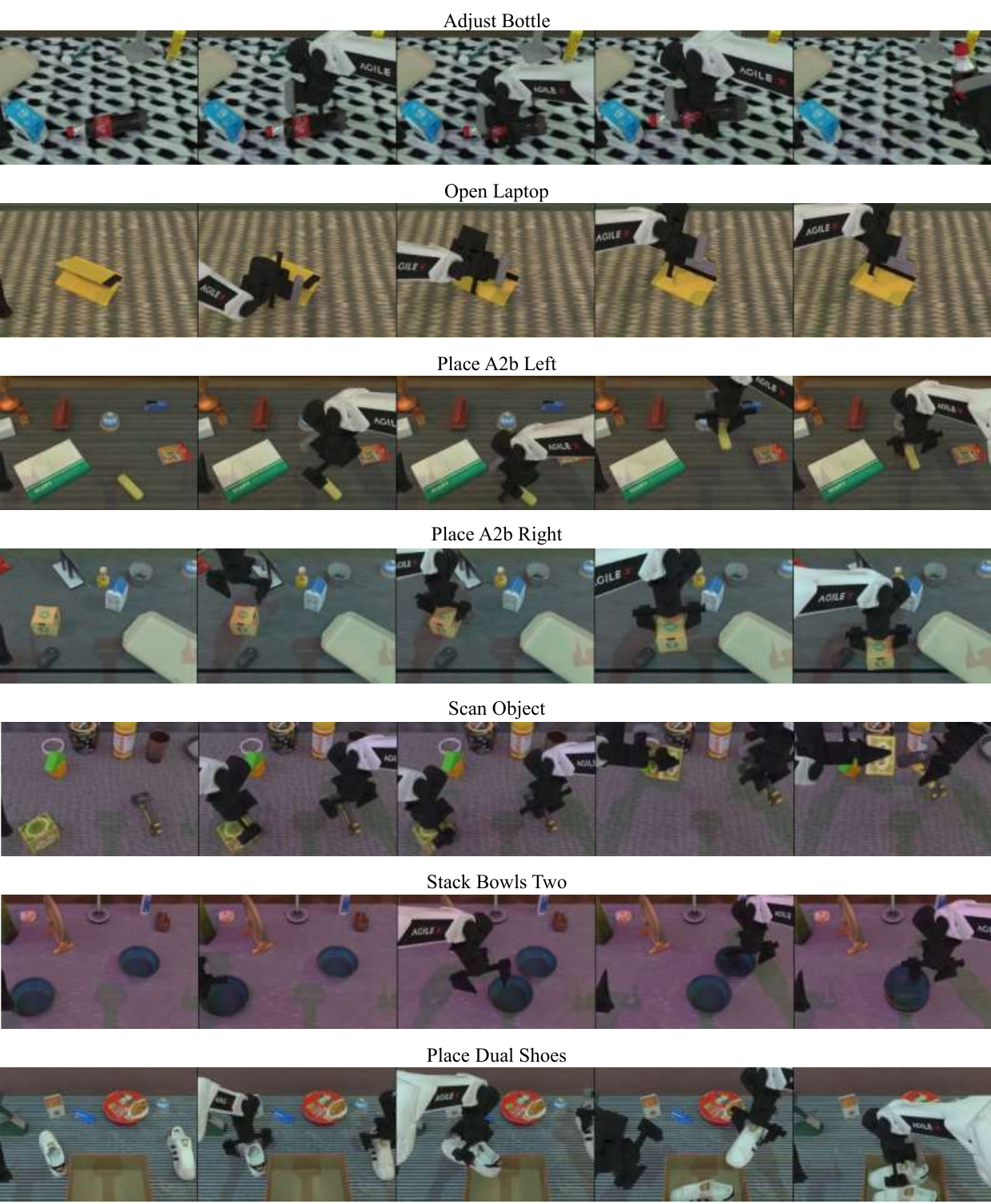}
    \caption{Qualitative examples on RoboTwin 2.0. Representative rollout snapshots, mainly under the Randomized environment, show that RoVLA maintains relatively stable target grounding and execution trajectories despite changes in object pose, scene layout, and visual appearance.}
    \label{fig:qualitative-robotwin2}
\end{figure*}

Figure~\ref{fig:qualitative-robotwin2} shows representative rollout examples on RoboTwin 2.0, with particular emphasis on the Randomized environment. Compared with the Clean setting, the randomized evaluation introduces joint changes in object pose, scene layout, and appearance, which makes policy execution more sensitive to grounding errors and temporal drift. Despite these perturbations, RoVLA generally preserves a more stable manipulation trajectory throughout the rollout. These qualitative observations are consistent with the quantitative improvements under randomization and further support that the proposed multi-consistency training enhances robust execution in complex manipulation environments.

\subsection{Qualitative analysis on real-world tasks}

\begin{figure*}[t]
    \centering
    \includegraphics[width=0.9\linewidth]{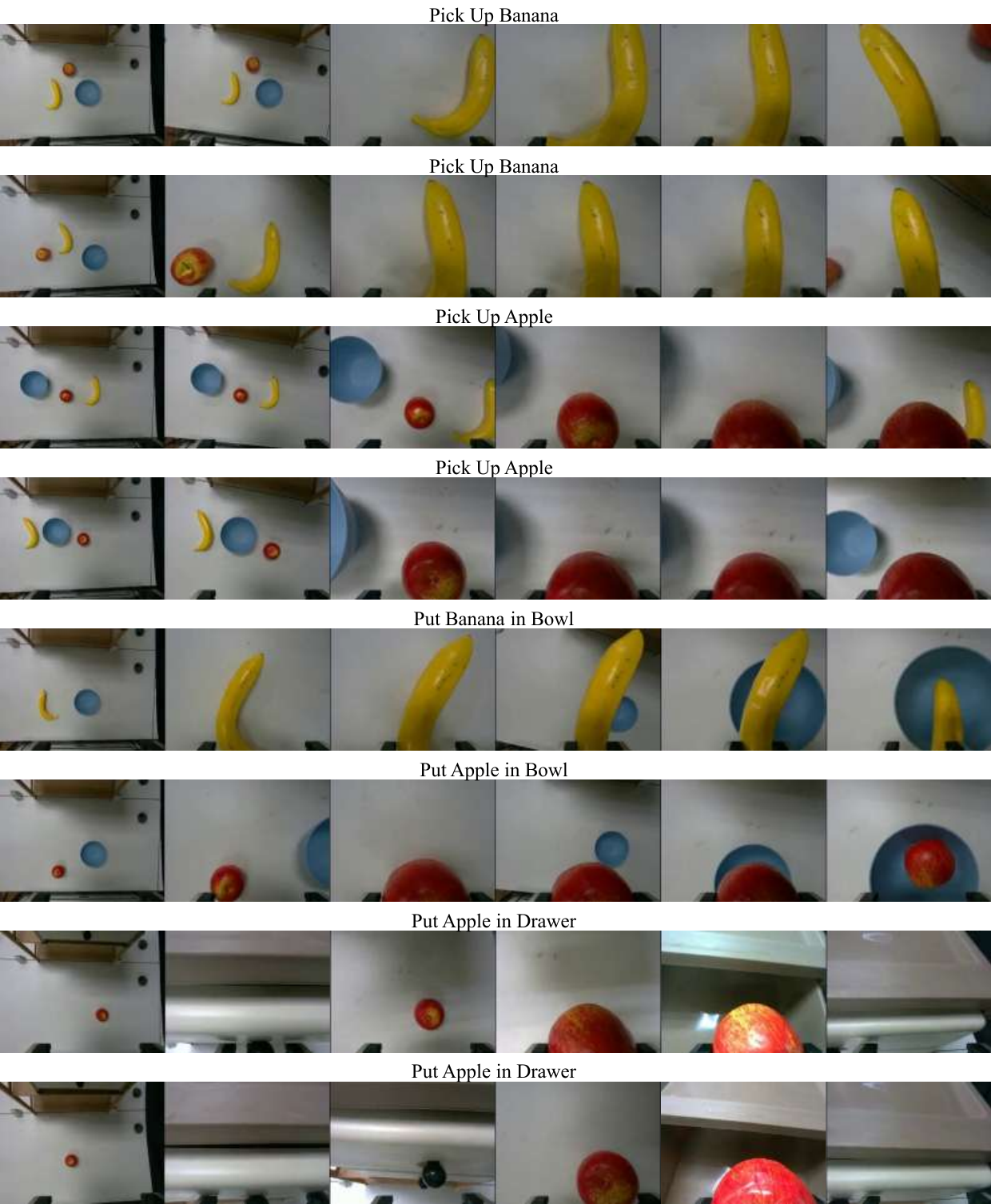}
    \caption{Qualitative examples on the real-world evaluation tasks visualized from the wrist-mounted camera view. Representative rollout snapshots from the five real-world tabletop manipulation tasks show that RoVLA maintains relatively stable execution trajectories even when the initial object configurations vary across trials.}
    \label{fig:qualitative-real_world_tasks}
\end{figure*}

Figure~\ref{fig:qualitative-real_world_tasks} presents representative rollout examples from the five real-world tabletop manipulation tasks introduced in the main paper. Although the task set is limited, these examples provide intuitive evidence that RoVLA can maintain relatively stable execution in real-world environments. In particular, even when the initial object configurations vary across trials, the policy still follows the intended manipulation procedure and produces comparatively coherent action trajectories. Notably, the bottom row illustrates a failure-retry behavior during drawer pulling: although the policy is trained exclusively on successful demonstrations, it can still respond to an out-of-distribution execution failure at test time and continue toward task completion. These qualitative results are consistent with the quantitative improvements reported in the main paper and further support the practical robustness of the proposed multi-consistency training strategy.

\end{document}